\documentclass[10pt,twocolumn,letterpaper]{article}

\usepackage{wacv}
\usepackage{times}
\usepackage{epsfig}
\usepackage{graphicx}
\usepackage{amsmath}
\usepackage{amssymb}

\usepackage[utf8]{inputenc} 
\usepackage[T1]{fontenc}    
\usepackage{url}            
\usepackage{booktabs}       
\usepackage{amsfonts}       
\usepackage{nicefrac}       
\usepackage{microtype}      
\usepackage{amsmath}
\usepackage{multirow}
\usepackage{subcaption}
\usepackage{cite}

\DeclareMathOperator{\Tr}{Tr}
\newcommand{\norm}[1]{\left\lVert#1\right\rVert}

\wacvfinalcopy

\ifwacvfinal
\usepackage[breaklinks=true,bookmarks=false]{hyperref}
\else
\usepackage[pagebackref=true,breaklinks=true,colorlinks,bookmarks=false]{hyperref}
\fi

\begin{document}

\title{InfoMax-GAN: Improved Adversarial Image Generation via \\ Information Maximization and Contrastive Learning}
\author{Kwot Sin Lee$^{1,2}$ \qquad Ngoc-Trung Tran$^3$ \qquad Ngai-Man Cheung$^3$
\\
$^1$University of Cambridge \qquad $^2$Snap Inc.\\
$^3$Singapore University of Technology and Design\\
}

\maketitle

\begin{abstract}
    While Generative Adversarial Networks (GANs) are fundamental to many generative modelling applications, they suffer from numerous issues. In this work, we propose a principled framework to simultaneously mitigate two fundamental issues in GANs: catastrophic forgetting of the discriminator and mode collapse of the generator. We achieve this by employing for GANs a contrastive learning and mutual information maximization approach, and perform extensive analyses to understand sources of improvements. Our approach significantly stabilizes GAN training and improves GAN performance for image synthesis across \textit{five} datasets under the \textit{same training and evaluation conditions} against state-of-the-art works. In particular, compared to the state-of-the-art SSGAN, our approach does not suffer from poorer performance on image domains such as faces, and instead improves performance significantly. Our approach is simple to implement and practical: it involves only one auxiliary objective, has a low computational cost, and performs robustly across a wide range of training settings and datasets \textit{without any hyperparameter tuning}. For reproducibility, our code is available in Mimicry \cite{lee2020mimicry}: \href{https://github.com/kwotsin/mimicry}{https://github.com/kwotsin/mimicry}.
\end{abstract}

\vspace{-0.6cm}
\section{Introduction}
The field of generative modelling has witnessed incredible successes since the advent of Generative Adversarial Networks (GANs) \cite{goodfellow2014generative}, a form of generative model known for its sampling efficiency in generating high-fidelity data \cite{odena2019open}. 
In its original formulation, a GAN is composed of two models - a generator and a discriminator - which together play an adversarial minimax game that enables the generator to model the true data distribution of some empirical data. This adversarial game is encapsulated by the following equation:
\begin{equation}
\begin{split}
    \min_{G}\max_{D} V(D, G) &=  
        \mathbb{E}_{x\sim p_r(x)}[\log D(x)] \\
        &+ \mathbb{E}_{z\sim p_(z)}[\log (1 - D(G(z)))] \\
\end{split}
\end{equation}
where $V$ is the value function, $p(z)$ is a prior noise distribution, $p_r(x)$ is the real data distribution, and $G(z)$ is the generated data from sampling some random noise $z$.

In this formulation, training the discriminator and generator with their respective minimax loss functions aims to minimize the Jensen-Shannon (JS) divergence between the real and generated data distributions \cite{goodfellow2014generative} $p_r$ and $p_g$ respectively. However, GAN training is notoriously difficult. Firstly, such theoretical guarantees only come under the assumption of the discriminator being trained to optimality \cite{gulrajani2017improved}, which may lead to saturating gradients in practice. Even so, there is no guarantee for convergence in this minimax game as both generator and discriminator are simultaneously and independently finding a Nash equilibrium in a high-dimensional space. Finally, GANs face the perennial problem of mode collapse, where $p_g$ collapses to only cover a few modes of $p_r$, resulting in generated samples of limited diversity. Consequently, recent years have seen concerted efforts \cite{odena2017conditional, tran2019improving, Tran_2018_ECCV, heusel2017gans, salimans2016improved, goodfellow2016nips, tran2019improved} to mitigate these issues.

A primary cause of GAN training instability is the non-stationary nature of the training environment: as the generator learns, the modeled distribution $p_g$ the discriminator faces is ever changing. As our GAN models are neural networks, the discriminator is susceptible to \textit{catastrophic forgetting} \cite{mccloskey1989catastrophic, chen2019self, kirkpatrick2017overcoming, kemker2018measuring}, a situation where the network learns ad-hoc representations and forgets about prior tasks to focus on the current one as the weights of the network updates, which contributes to training instability. The state-of-the-art Self-supervised GAN (SSGAN) \cite{chen2019self} is the first to demonstrate that a representation learning approach could mitigate discriminator catastrophic forgetting, thus improving training stability. However, the approach still does not explicitly mitigate \textit{mode collapse}, and has a failure mode in image domains such as faces \cite{chen2019self}. Furthermore, \cite{tran2019self} shows that while SSGAN's approach is helpful for discriminator forgetting, it in fact promotes mode collapse for the generator.

To overcome these problems, we present an approach to simultaneously mitigate both catastrophic forgetting and mode collapse. On the discriminator side, we apply mutual information maximization to improve long-term representation learning, thereby reducing catastrophic forgetting in the non-stationary training environment. On the generator side, we employ contrastive learning to incentivize the generator to produce diverse images that give easily distinguishable positive/negative samples, hence reducing mode collapse.
Through mitigating both issues, we show a wide range of practical improvements on natural image synthesis using GANs.
We summarize our contributions below:
\begin{itemize}
    \item We present a GAN framework to improve natural image synthesis through simultaneously mitigating two key GAN issues using just one objective: catastrophic forgetting of the discriminator (via information maximization) and mode collapse of the generator (via contrastive learning). Our approach mitigates issues in both discriminator and generator, rather than either alone.
    
    \item With this multi-faceted approach, we significantly improve GAN image synthesis across \textit{five} different datasets against state-of-the-art works under the \textit{same training and evaluation conditions}.
    
    \item Our framework is lightweight and practical: it introduces just one auxiliary objective, has a low computational cost, and is robust against a wide range of training settings \textit{without any tuning} required.
    
    \item Our work is the first to demonstrate the effectiveness of contrastive learning for significantly improving GAN performance, and also the first to apply the InfoMax principle in a GAN setting, which we hope would open a new research direction in these areas.
\end{itemize}

\vspace{-0.2cm}
\section{Background}
\paragraph{Mutual information and representation learning}
Mutual information has deep connections with representation learning \cite{bengio2013representation}, where we aim to learn an encoder function $E$ that ideally captures the most important features of the input data $X$, often at a lower dimensional latent space. This concept is encapsulated by the InfoMax objective \cite{linsker1988self}:
\vspace{-0.2cm}
\begin{equation}
    \max_{E \in \mathcal{E}} \mathcal{I}(X; E(X))
    \vspace{-0.2cm}
\end{equation}
where $\mathcal{E}$ is some function class, and the objective is to find some $E$ that maximizes the mutual information between the input data and its encoded representations $E(X)$. To maximize on the InfoMax objective, one could alternatively maximize $\mathcal{I}(C_\psi(X); E_\psi(X))$, where $C_\psi$ and $E_\psi$ are encoders part of the same architecture parameterised by $\psi$.
It is shown in \cite{tschannen2019mutual} maximizing $\mathcal{I}(C_\psi(X); E_\psi(X))$ is maximizing on a lower bound of the InfoMax objective:
\vspace{-0.2cm}
\begin{equation}
    \mathcal{I}(C_\psi(X); E_\psi(X)) \leq \mathcal{I}(X; (C_\psi(X), E_\psi(X)))
    \label{lower_bound}
    \vspace{-0.2cm}
\end{equation}

In practice, maximizing $\mathcal{I}(C_\psi(X); E_\psi(X))$ has several advantages: (a) Using different feature encodings allow us to capture different views and modalities of the data for flexibility of modelling \cite{bachman2019learning, tian2019contrastive};
(b) The encoded data lies in a much lower dimensional latent space than that of the original data, thus reducing computational constraints \cite{tschannen2019mutual, poole2018variational}.

\paragraph{Contrastive learning} Recently, state-of-the-art works in unsupervised representation learning \cite{bachman2019learning, oord2018representation, hjelm2018learning, henaff2019data, tian2019contrastive, lowe2019greedy, kong2019mutual} lies in taking a contrastive approach to maximizing the mutual information between encoded local and global features. Yet, since directly maximizing mutual information is often intractable in practice \cite{paninski2003estimation}, these works often maximize on the InfoNCE \cite{oord2018representation} lower bound instead, which involves a contrastive loss minimized through having a critic find positive samples in contrast to a set of negative samples. Such positive/negative samples are arbitrarily created by pairing features \cite{hjelm2018learning}, augmentation \cite{chen2020simple}, or a combination of both \cite{bachman2019learning}. Our work similarly maximizes on this InfoNCE bound, and most closely follows the Deep InfoMax \cite{hjelm2018learning} approach of obtaining local and global features for the maximization.
\vspace{-0.3cm}
\section{InfoMax-GAN}

\begin{figure*}
\centering
  \includegraphics[width=0.77\linewidth]{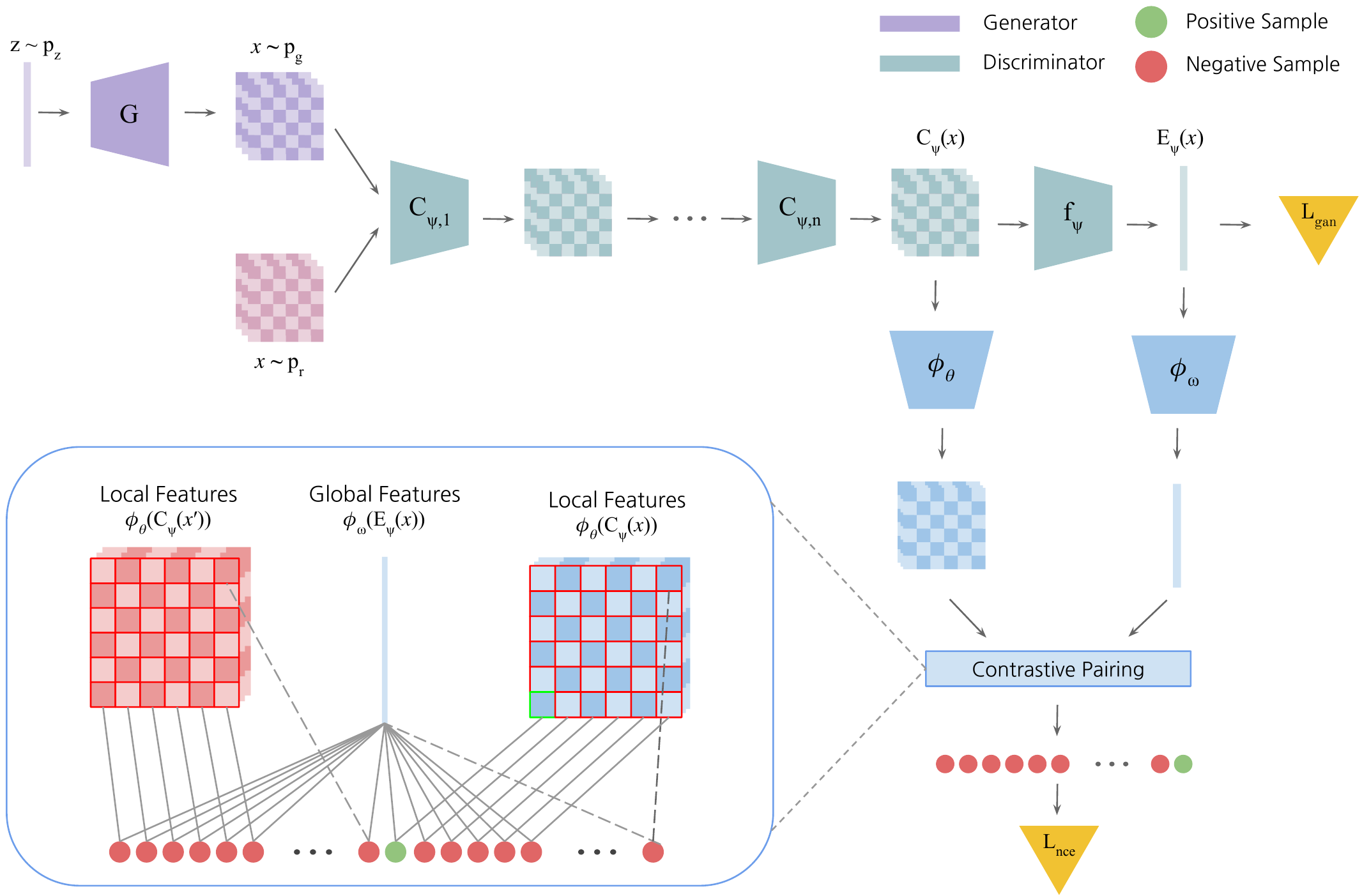}
  \caption{Illustration of the InfoMax-GAN framework. An image $x$ is sampled from the real data distribution $p_r$ or fake data distribution $p_g$ as modeled by the generator $G$. Image $x$ passes through a discriminator encoder $E_\psi = f_\psi \circ C_\psi$, where $C_\psi = C_{\psi, 1} \circ ... \circ C_{\psi, n}$ is a series of $n$ intermediate discriminator layers leading to the last local feature map $C_\psi(x)$ and $f_\psi$ transforms $C_\psi(x)$ to a global feature vector $E_\psi(x)$, which is subsequently used to compute the GAN objective $\mathcal{L}_{\text{gan}}$. The local and global features $C_\psi(x)$ and $E_\psi(x)$ are then projected to a higher dimension by the spectral normalized critic networks $\phi_\theta$ and $\phi_\omega$ respectively. Finally, the resulting features undergo a Contrastive Pairing phase involving local features from another image $x'$, to produce positive and negative samples for computing the contrastive loss $\mathcal{L}_{\text{nce}}$.}
  \label{framework}
  \vspace{-0.5cm}
\end{figure*}

\subsection{Approach}
Figure \ref{framework} illustrates the InfoMax-GAN framework.
Firstly, to maximize on the lower bound of the InfoMax objective, $\mathcal{I}(C_\psi(X); E_\psi(X))$, we set $E_\psi$ to represent layers of the discriminator leading to the global features, and $C_\psi$ as layers leading to the local features. Here, $C_\psi = C_{\psi, 1} \circ ... \circ C_{\psi, n}$ is a series of $n$ intermediate discriminator layers leading to the last local feature map $C_\psi(x)$ and $f_\psi$ is the subsequent layer transforming $C_\psi(x)$ to a global feature vector $E_\psi(x)$, which is ultimately used for computing the GAN objective $\mathcal{L}_{\text{gan}}$. We set the local and global feature as the penultimate and final feature outputs of the discriminator encoder respectively, and we study its ablation impact in Appendix \ref{appendix: ablation}.

Next, the local/global features $C_\psi(x)$ and $E_\psi(x)$ extracted from the discriminator are passed to the critic networks $\phi_\theta$ and $\phi_\omega$ to be projected to a higher dimension Reproducing Kernel Hilbert Space (RKHS) \cite{aronszajn1950theory}, which exploits the value of linear evaluation in capturing similarities between the global and local features. These projected features then undergo a Contrastive Pairing phase to create positive/negative samples, where given some image $x$, a positive sample is created by pairing the (projected) global feature vector $\phi_\omega(E_\psi(x))$ with a (projected) local spatial vector $\phi_\theta(C_\psi^{(i)}(x))$ from the image's own (projected) local feature map $\phi_\theta(C_\psi(x))$, where $i \in \mathcal{A} = \{0, 1, ..., M^2-1\}$ is an index to the $M \times M$ local feature map. Doing so, we represent a positive sample as the pair $(\phi_\theta(C_\psi^{(i)}(x)), \phi_\omega(E_\psi(x)))$ for some $i$. For each of such positive sample, negative samples are obtained by sampling local spatial vectors from the projected local feature map of another image $x'$ in the same mini-batch, and are represented as the pairs $(\phi_\theta(C_\psi^{(i)}(x')), \phi_\omega(E_\psi(x)))$. Intuitively, this step constrains the discriminator to produce global features of some image that maximizes mutual information with the local features of the same image, rather than those from other images.

Taking this further, consider for each positive sample, the pairs $(\phi_\theta(C_\psi^{(j)}(x)), \phi_\omega(E_\psi(x)))$, $j \in \mathcal{A}, j \neq i$ as negative samples. That is, using spatial vectors from the \textit{same} local feature map to create negative samples. Doing so, we regularize the learnt representations to avoid trivial solutions to the mutual information maximization objective, since the global features are constrained to have consistently high mutual information with \textit{all} spatial vectors of its own local feature map, rather than from only some. This effectively aggregates all local 
information of the image to represent it.

Thus, for $N$ images in a mini-batch, we produce positive/negative samples to perform an $NM^2$ way classification for each positive sample. Through this approach, it is shown in \cite{oord2018representation} one maximizes the InfoNCE lower bound of the mutual information $\mathcal{I}(C_\psi(X); E_\psi(X))$. 
Formally, for a set of N random images $X = \{x_1, ..., x_N\}$ and set $\mathcal{A} = \{0, 1, ..., M^2-1\}$ representing indices of a $M \times M$ spatial sized local feature map, the contrastive loss is:
\begin{equation}
\label{infonce_eqn}
\begin{split}
    \mathcal{L}_{nce}(X)
    &= -\mathbb{E}_{x \in X} \mathbb{E}_{i \in \mathcal{A}}
    \left[
    \log
    p(C_\psi^{(i)}(x),  E_\psi(x) \mid X)
    \right] \\
    &= -\mathbb{E}_{x \in X} \mathbb{E}_{i \in \mathcal{A}}
    \left[
    \Delta
    \right] \\
    \Delta &= \log
    {
    \frac
    {\exp(g_{\theta, \omega}(C_\psi^{(i)}(x), E_\psi(x)))}
    {\sum_{(x', i) \in X \times \mathcal{A}} \exp(g_{\theta, \omega}(C_\psi^{(i)}(x'), E_\psi(x)))}
    }
\end{split}
\end{equation}
where $g_{\theta, \omega}: \mathbb{R}^{1 \times 1 \times K} \times \mathbb{R}^{1 \times 1 \times K} \rightarrow \mathbb{R}$ is a critic mapping the local/global features with $K$ dimensions to a scalar score.
Formally, we define $g_{\theta, \omega}$ to be:
\begin{equation}
    g_{\theta, \omega}(C_\psi^{(i)}(x), E_\psi(x)) = \phi_\theta(C_\psi^{(i)}(x))^T \phi_\omega(E_\psi(x))
\end{equation}
where $\phi_\theta: \mathbb{R}^{M \times M \times K} \rightarrow \mathbb{R}^{M \times M \times R}$ and $\phi_\omega: \mathbb{R}^{1 \times 1 \times K} \rightarrow \mathbb{R}^{1 \times 1 \times R}$ are the critic networks parameterized by $\theta$ and $\omega$ respectively, projecting the local and global features to the higher RKHS. In practice, $\phi_\theta$ and $\phi_\omega$ are defined as shallow networks with only 1 hidden layer following \cite{hjelm2018learning}, but with spectral normalized weights as well. These shallow networks serve to only project the feature dimensions of the input features, and preserve their original spatial sizes.

To stabilize training, we constrain the discriminator to learn from only the contrastive loss of real image features, and similarly for the generator, from only the contrastive loss of fake image features. We formulate the losses for discriminator and generator $\mathcal{L}_D$ and $\mathcal{L}_G$ as such:
\begin{equation}
    \mathcal{L}_G = \mathcal{L}_{\text{gan}}(\hat D, G) + \alpha \mathcal{L}_{\text{nce}}(X_g)
\end{equation}
\begin{equation}
    \mathcal{L}_D = \mathcal{L}_{\text{gan}}(D, \hat G) + \beta \mathcal{L}_{\text{nce}}(X_r)
\end{equation}
where $\alpha$ and $\beta$ are hyperparameters; $\hat D$ and $\hat G$ represent a fixed discriminator and generator respectively; $X_r$ and $X_g$ represent sets of real and generated images respectively; and $\mathcal{L}_{\text{gan}}$ is the hinge loss for GANs \cite{miyato2018spectral}:
\begin{equation}
    \begin{split}
            \mathcal{L}_{\text{gan}}(D, \hat G) &= \mathbb{E}_{x\sim p_r} [\min(0, 1 - D(x))] \\
            &+ \mathbb{E}_{z \sim p_z}[\min(0, 1 + D(\hat{G}(z)))]
    \end{split}
\end{equation}
\vspace{-0.25cm}
\begin{equation}
    \mathcal{L}_{\text{gan}}(\hat D, G) = - \mathbb{E}_{z \sim p_z} [\hat{D}(G(z))]
\end{equation}
In practice, we set $\alpha = \beta = 0.2$ for all experiments for simplicity, with ablation studies to show our approach is robust across a wide range of $\alpha$ and $\beta$ values.

\begin{figure}
\centering
  \includegraphics[width=0.7\linewidth]{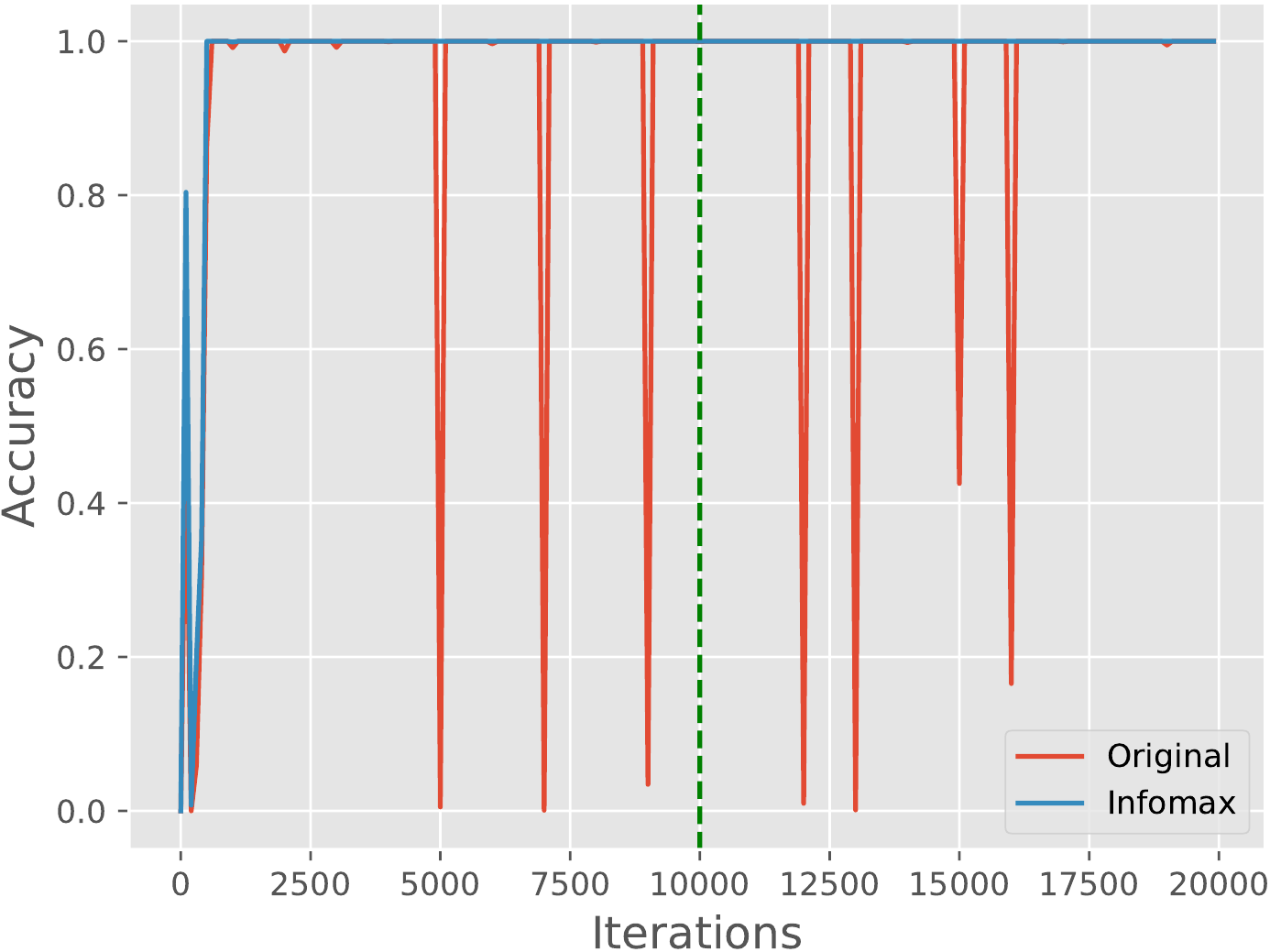}
  \caption{Accuracy of a classifier when trained on the one-vs-all CIFAR-10 classification task. Regularized with the InfoMax objective by minimizing (\ref{infonce_eqn}), the classifier successfully predicts classes trained from previous iterations even when the underlying class distribution changes.} 
  \label{catastrophic_curves}
  \vspace{-0.5cm}
\end{figure}

\subsection{Mitigating Catastrophic Forgetting}
Our approach mitigates a key issue in GANs: \textit{catastrophic forgetting} of the discriminator, a situation where due to the non-stationary GAN training environment, the discriminator learns only ad-hoc representations and forget about prior tasks it was trained on. For instance, while the discriminator may learn to penalize flaws in global structures early in GAN training \cite{chen2019self}, it may later forget these relevant representations in order to learn those for finding detailed flaws in local structures, which overall contributes to training instability.

Inspired by \cite{chen2019self}, we examine the ability of our approach in mitigating catastrophic forgetting: we train a discriminator classifier on the one-vs-all CIFAR-10 classification task where the underlying class distribution changes every 1K iterations, and the cycle repeats every 10K iterations. As seen in Figure \ref{catastrophic_curves}, without the InfoMax objective, the classifier overfits to a certain class distribution and produces very low accuracy when the class distribution is changed. When training is regularized with the InfoMax objective, the classifier successfully remembers all prior classes it was trained on. Thus, the InfoMax objective helps the discriminator to reduce catastrophic forgetting and adapt to the non-stationary nature of the generated image distribution, which ultimately stabilizes GAN training.

\begin{figure}
\centering
\begin{subfigure}{0.35\textwidth}
  \centering
    \includegraphics[width=\linewidth]{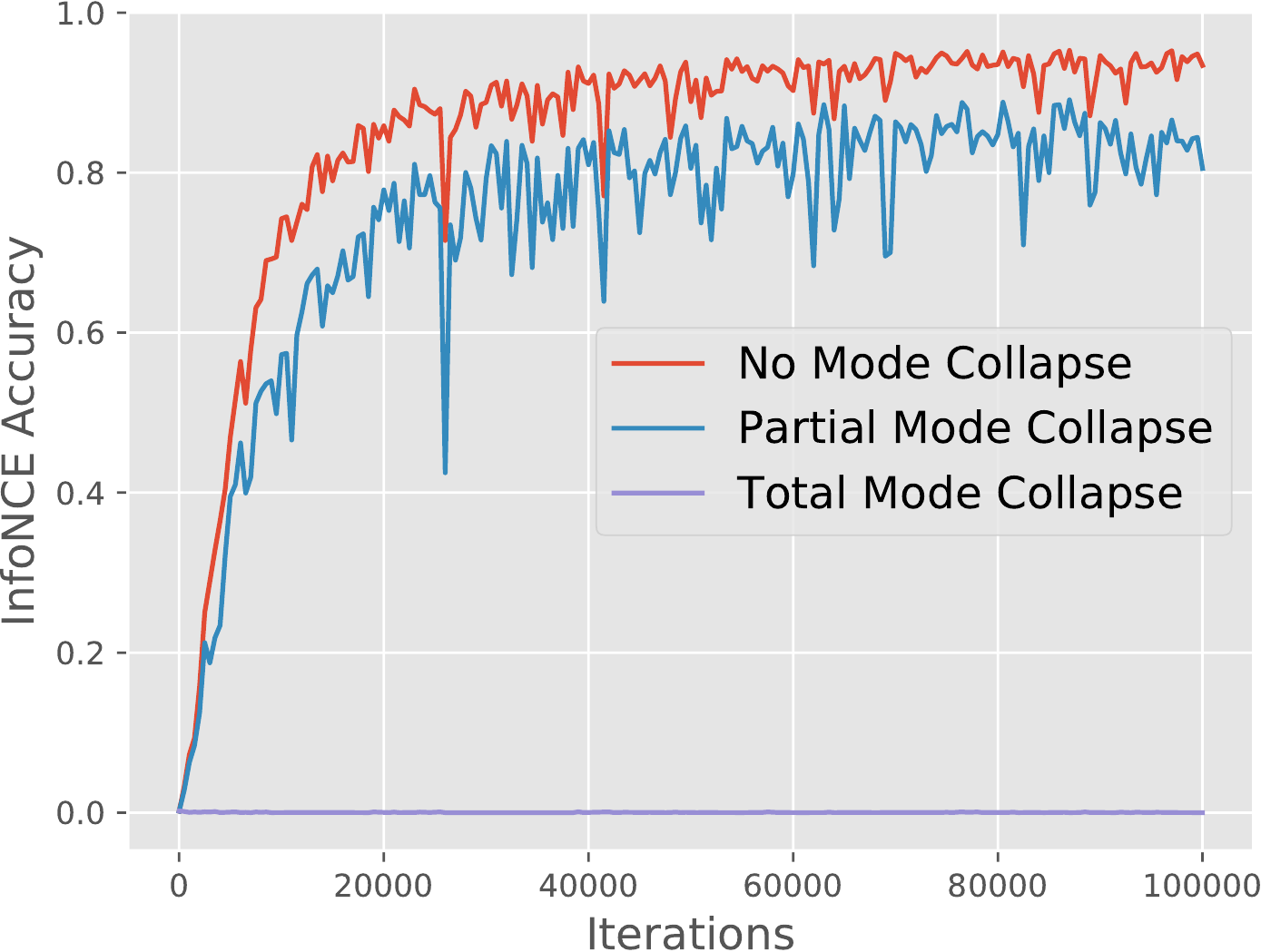}
    \caption{}
  \label{mode_collapse:a}
\end{subfigure}
\begin{subfigure}{0.35\textwidth}
  \centering
    \includegraphics[width=\linewidth]{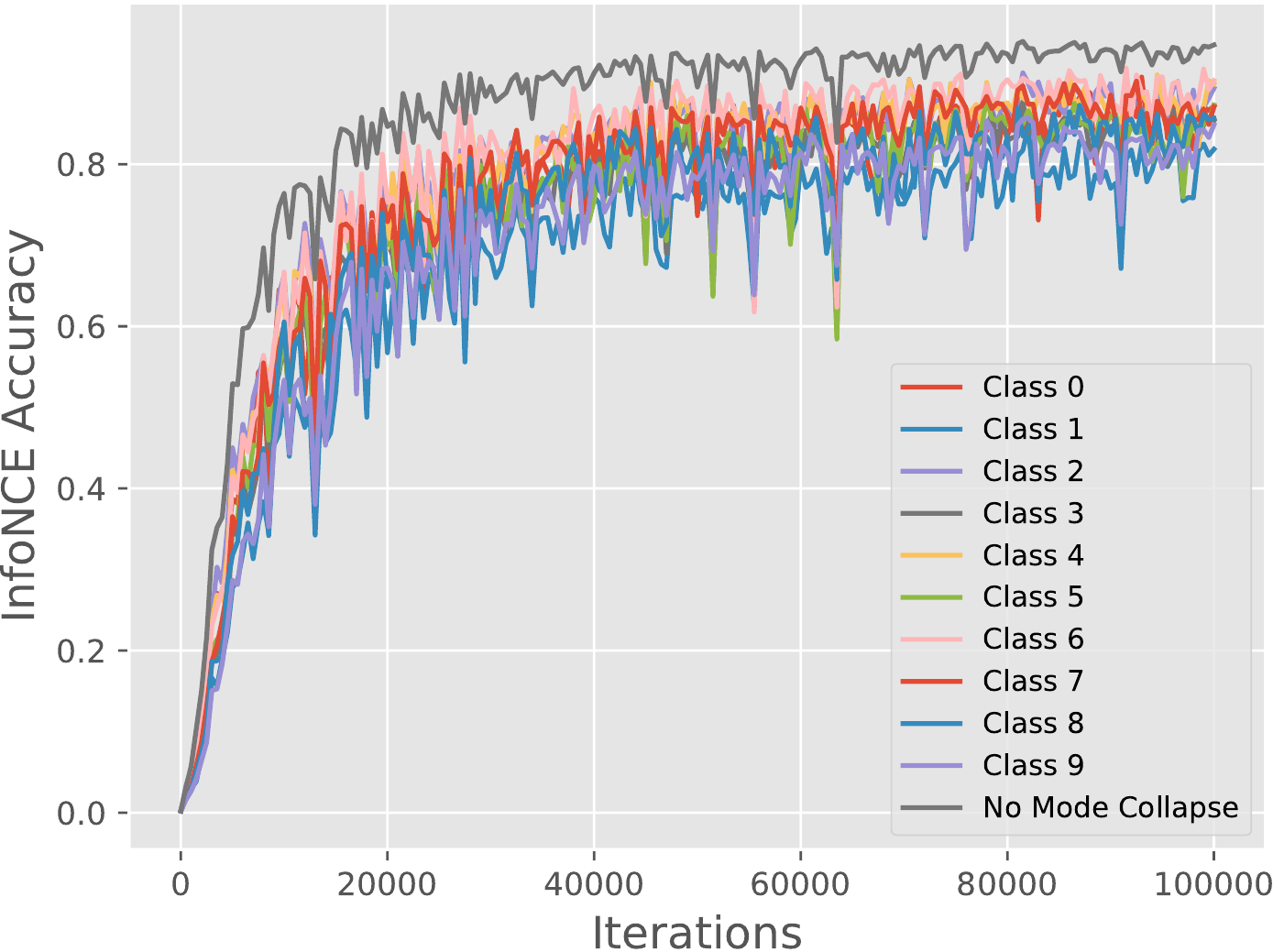}
  \caption{}
  \label{mode_collapse:b}
\end{subfigure}
\caption{Contrastive task accuracy when simulating generators exhibiting a range of mode collapse behaviours using CIFAR-10 data. (a) We show that the less mode collapsed a generator is, the better the accuracy for contrastive task. (b) The contrastive task accuracy is consistently lower when the generator has partially mode collapsed to any individual class, compared to when there is no mode collapse.}
\label{mode_collapse_curve}
\vspace{-0.5cm}
\end{figure}

\subsection{Mitigating Mode Collapse}
Our approach also mitigates a persistent problem of the generator: \textit{mode collapse}. For a fully mode collapsed generator, we have 
$ x = x'$ $\forall x,x' \sim X_g$, where $X_g$ is a set of randomly generated images, such that $C_\psi(x) = C_\psi(x')$.
This means the term $p(C_\psi^{(i)}(x),  E_\psi(x) \mid X_g)$ approaches 0 in the limit, rather than the optimal value $1$, as the critics cannot distinguish apart the multiple identical feature pairs from individual images.

To validate this, we show there is a direct correlation between the diversity of generated images and the contrastive learning task accuracy $p(C_\psi^{(i)}(x),  E_\psi(x) \mid X)$. We train the discriminator to solve the contrastive task using CIFAR-10 training data, and simulate 3 different kinds of generators using CIFAR-10 test data: (a) a perfect generator with no mode collapse that generates \textit{all classes of images}; (b) a partially mode collapsed generator that only generates \textit{one class of images} and (c) a totally mode collapsed generator that only generates \textit{one image}. 

From Figure \ref{mode_collapse:a}, we observe a perfect generator with no mode collapse best solves the contrastive task, and a partially mode collapsed generator has a consistently poorer accuracy in the contrastive task than the perfect generator.  This concurs with our expectation: images from only one class exhibit a much lower diversity than images from all classes, and so distinguishing positive samples amongst similar and harder negative samples makes the contrastive task much harder. Furthermore, for a totally mode collapsed generator which only generates one image, the accuracy is \textit{near zero}, which confirms our initial hypothesis. For any $N$ images, there are $NM^2$ samples to classify in the contrastive task, with $NM^2 - 1$ negative samples for each positive sample. However, if all $N$ images are identical due to total mode collapse, then there exists $N-1$ negative samples identical to each positive sample, which makes the contrastive task nearly impossible to solve. Thus, to solve the contrastive task well, the generator is highly encouraged to generate images with greater diversity, which reduces mode collapse. 

Furthermore, in Figure \ref{mode_collapse:b}, the performance of any class demonstrating partial mode collapse is \textit{consistently worse} than the case of no mode collapse, where all classes of images are used. Thus, the generator is incentivized to not collapse to producing just any one class that fools the discriminator easily, since producing \textit{all classes} of images naturally leads to the best performance in the contrastive task.

\section{Experiments}
\subsection{Experimental Settings}
\paragraph{GAN architectures} We compare our model with the baseline Spectral Normalization GAN (SNGAN) \cite{miyato2018spectral} and the state-of-the-art Self-supervised GAN (SSGAN) \cite{chen2019self}. For clarity, we highlight InfoMax-GAN is equivalent to SNGAN with our proposed objective, and SSGAN is equivalent to SNGAN with the rotation task objective. We show InfoMax-GAN \textit{alone} performs highly competitively, with significant improvements over SSGAN. We detail the exact architectures used for all models and datasets in Appendix \ref{appendix:B}.

\paragraph{Datasets} We experiment on five different datasets at multiple resolutions:
ImageNet ($128 \times 128$) \cite{deng2009imagenet}, CelebA ($128 \times 128)$ \cite{liu2015faceattributes}, CIFAR-10 ($32 \times 32$) \cite{krizhevsky2009learning}, STL-10 ($48 \times 48$) \cite{coates2011analysis}, and CIFAR-100 ($32 \times 32$) \cite{krizhevsky2009learning}. The details for these datasets can be found in Appendix \ref{appendix: dataset_settings}.

\paragraph{Training}  We train all models using the same Residual Network \cite{he2016deep} backbone, under the \textit{exact} same settings for each dataset, and using the same code base, for fairness in comparisons. For details, refer to Appendix \ref{appendix: training_settings}. For all models and datasets, we set $\alpha=\beta=0.2$, to balance the contrastive loss to be on the same scale as the GAN loss initially. This scaling principle is similar to what is applied in \cite{chen2016infogan}, and we later show in our ablation study our framework is highly robust to changes in these hyperparameters.

\paragraph{Evaluation} 
To assess the generated images quality, we employ three different metrics: Fréchet Inception Distance (FID) \cite{heusel2017gans}, Kernel Inception Distance (KID) \cite{binkowski2018demystifying}, and Inception Score (IS) \cite{salimans2016improved}. In general, FID and KID measure the diversity of generated images, and IS measures the quality of generated images. Here, we emphasize we use the \textit{exact} same number of real and fake samples for evaluation, so that we can compare the scores fairly. This is crucial, especially since metrics like FID can produce highly biased estimates \cite{binkowski2018demystifying}, where using a larger sample size leads to a significantly lower score. Finally, for all our scores, we compute them using 3 different random seeds to report their mean and standard deviation. A detailed explanation of all three metrics and the sample sizes used can be found in Appendix \ref{appendix: eval_metrics}

\subsection{Results}
\begin{table*}
\centering
\scalebox{0.95}{
\begin{tabular}{c|*{6}c}
\toprule
\multirow{2}{*}{\textbf{Metric}} 
& \multirow{2}{*}{\textbf{Dataset}} 
& \multirow{2}{*}{\textbf{Resolution}} 
& \multicolumn{3}{c}{\textbf{Models}} 
\\ \cmidrule(l){4-6} 
&  &  
& \textbf{SNGAN} 
& \textbf{SSGAN} 
& \textbf{InfoMax-GAN}
\\ \midrule
\multirow{7}{*}{FID} 
    & ImageNet 
        & $128 \times 128$ 
        & $65.74 \pm 0.31$
        & $62.48 \pm 0.31$
        & $\mathbf{58.91 \pm 0.14}$ \\
    & CelebA 
        & $128 \times 128$
        & $14.04 \pm 0.02$ 
        & $16.39 \pm 0.09$ 
        & $\mathbf{10.63 \pm 0.04}$ \\
    & STL-10 & $48 \times 48$
        & $40.48 \pm 0.07$
        & $38.97 \pm 0.23$
        & $\mathbf{37.49 \pm 0.05}$ \\
    & CIFAR-100 
        & $32 \times 32$
        & $24.76 \pm 0.16$ 
        & $24.64 \pm 0.16$
        & {$\mathbf{21.22 \pm 0.26}$} \\
    & CIFAR-10 
        & $32 \times 32$
        & $18.63 \pm 0.22$
        & $\mathbf{16.59 \pm 0.13}$
        & $17.14 \pm 0.20$ \\
        
\midrule
\multirow{7}{*}{KID} 
 & ImageNet & $128 \times 128$ & $0.0663 \pm 0.0004$ & $0.0616 \pm 0.0004$ & $\mathbf{0.0579 \pm 0.0004}$\\
 & CelebA & $128 \times 128$ & $0.0076 \pm 0.0001$ & $0.0101 \pm 0.0001$ & $\mathbf{0.0063 \pm 0.0001}$ \\
 & STL-10 & $48 \times 48$ & $0.0369 \pm 0.0002$ & $0.0332 \pm 0.0004$ & $\mathbf{0.0326 \pm 0.0002}$ \\
 & CIFAR-100 & $32 \times 32$ & $0.0156 \pm 0.0003$ & $0.0161 \pm 0.0002$ & $\mathbf{0.0135 \pm 0.0004}$ \\
 & CIFAR-10 & $32 \times 32$ & $0.0125 \pm 0.0001$ & $\mathbf{0.0101 \pm 0.0002}$ & $0.0112 \pm 0.0001$ \\
\midrule
\multirow{7}{*}{IS} 
& ImageNet & $128 \times 128$ & $13.05 \pm 0.05
$ & $13.30 \pm 0.03$ & $\mathbf{13.68 \pm 0.06}$ \\
 & CelebA & $128 \times 128$ & $2.72 \pm 0.01$ & $2.63 \pm 0.01$ & $\mathbf{2.84 \pm 0.01}$ \\
 & STL-10 & $48 \times 48$ & $8.04 \pm 0.07$ & $8.25 \pm 0.06$ & $\mathbf{8.54 \pm 0.12}$ \\
 & CIFAR-100 & $32 \times 32$ & $7.57 \pm 0.11$ & $7.56 \pm 0.07$ & $\mathbf{7.86 \pm 0.10}$ \\
 & CIFAR-10 & $32 \times 32$ & $7.97 \pm 0.06$ & $\mathbf{8.17 \pm 0.06}$ & $8.08 \pm 0.08$ \\        
\bottomrule
\end{tabular}
}
\vspace{0.1cm}
\caption{Mean FID, KID and IS scores of all models across different datasets, computed across 3 different seeds. FID and KID: lower is better. IS: higher is better.}
\label{main_fid}
\vspace{-0.5cm}
\end{table*}
\paragraph{Improved image synthesis}
As seen in Table \ref{main_fid}, InfoMax-GAN improves FID consistently and significantly across many datasets over SNGAN and SSGAN. On the challenging high resolution ImageNet dataset, InfoMax-GAN improves by \textbf{6.8 points} over SNGAN, and \textbf{3.6 points} over SSGAN. On the high resolution CelebA, while SSGAN could not improve over the baseline SNGAN, as similarly noted in \cite{chen2019self}, InfoMax-GAN improves by \textbf{3.4 points} over SNGAN, and \textbf{5.8 points} over SSGAN. This suggests our approach is versatile and can generalise across multiple data domains.

On STL-10, InfoMax-GAN achieves an improvement of \textbf{3.0 points} over SNGAN and \textbf{1.5 points} over SSGAN. Interestingly, while InfoMax-GAN performs similarly as SSGAN on CIFAR-10 with around \textbf{0.5 points} difference, it improves FID by \textbf{3.4 points} on CIFAR-100 when the number of classes increase. We conjecture this is due to the tendency for SSGAN to generate easily rotated images \cite{tran2019self}, which sacrifices diversity and reduces FID when there are more classes. This observation also supports InfoMax-GAN's larger improvements on ImageNet, which has 1000 classes.

Similarly, for alternative metrics like KID and IS, InfoMax-GAN achieves a highly competitive performance and improves over the state-of-the-art works. On IS, InfoMax-GAN improves from \textbf{0.2 to 0.4 points} over SSGAN for all datasets except CIFAR-10, where the margin is less than \textbf{0.1 points} and within the standard deviation, indicating a similar performance. Similar to its FID performance on CelebA, SSGAN also performs worse in terms of IS compared to the baseline SNGAN, suggesting its failure mode on faces is not just due to a limited diversity, but also due to poorer quality. In contrast, InfoMax-GAN improves on IS over SNGAN and SSGAN significantly. Finally, on KID, we confirm our result on FID: where FID is better, KID is also better. This further substantiates our FID results and how InfoMax-GAN generates more diverse images across these datasets, with no obvious failure modes unlike in SSGAN.

\begin{table}[]
\centering
\scalebox{0.95}{
\begin{tabular}{@{}cccc@{}}
\toprule
\multirow{2}{*}{\textbf{Dataset}} & \multirow{2}{*}{\textbf{Resolution}} & \multicolumn{2}{c}{\textbf{Models}} \\ \cmidrule(l){3-4} 
 &  & \textbf{SSGAN} & \textbf{SSGAN + IM} \\ \midrule
ImageNet & $128 \times 128$ & $62.48 \pm 0.31$ & $\mathbf{56.45 \pm 0.29}$ \\
CelebA & $128 \times 128$ & $16.39 \pm 0.09$ & $\mathbf{11.93 \pm 0.14}$ \\
STL-10 & $48 \times 48$ & $38.97 \pm 0.23$ & $\mathbf{37.73 \pm 0.06}$ \\
CIFAR-100 & $32 \times 32$ & $24.64 \pm 0.16$ & $\mathbf{21.40 \pm 0.20}$ \\
CIFAR-10 & $32 \times 32$ & $16.59 \pm 0.13$ & $\mathbf{15.42 \pm 0.08}$ \\
\bottomrule
\end{tabular}
}
\vspace{0.1cm}
\caption{Mean FID scores (lower is better) of SSGAN before and after applying our method: ``+ IM'' refers to adding our proposed InfoMax-GAN objective.}
\label{tab:ssgan_ortho}
\vspace{-0.5cm}
\end{table}
\paragraph{Orthogonal improvements} In Table \ref{tab:ssgan_ortho}, we show our improvements are orthogonal to those in SSGAN: when adding our objective into SSGAN, FID improves across all datasets significantly, achieving even larger improvements of approximately \textbf{2.5 points} for the challenging ImageNet dataset. Thus, our method is flexible and can be easily integrated into existing state-of-the-art works like SSGAN.

\begin{table*}
\centering
\scalebox{0.95}{
\begin{tabular}{cccccccccccccccccccccc}
\toprule
    \multirow{2}{*}{\textbf{$\beta_1$}}
    & \multirow{2}{*}{\textbf{$\beta_2$}}
    & \multirow{2}{*}{\textbf{$n_{\text{dis}}$}}
    & \multicolumn{2}{c}{\textbf{CIFAR-10}}
    & \multicolumn{2}{c}{\textbf{STL-10}}
    \\
    \cmidrule(lr){4-5}
    \cmidrule(lr){6-7}
    & 
    \multicolumn{1}{c}{} & \multicolumn{1}{c}{}
    & 
    \multicolumn{1}{c}{\textbf{SNGAN}} & \multicolumn{1}{c}{\textbf{InfoMax-GAN}}
    &
    \multicolumn{1}{c}{\textbf{SNGAN}} & \multicolumn{1}{c}{\textbf{InfoMax-GAN}}
    \\
\midrule
    0.0 & 0.9 & 1
    & $164.74 \pm 0.42$ & $\mathbf{24.42 \pm 0.18}$
    & $267.10 \pm 0.20$ & $\mathbf{54.29 \pm 0.13}$
    \\
    0.0 & 0.9 & 2
    & $20.87 \pm 0.19$ & $\mathbf{18.08 \pm 0.27}$
    & $46.65 \pm 0.18$ & $\mathbf{38.96 \pm 0.31}$
    \\
    0.0 & 0.9 & 5 
    & $18.63 \pm 0.22$ & {$\mathbf{17.14 \pm 0.20}$}
    & $40.48 \pm 0.07$ & $\mathbf{37.49 \pm 0.05}$
    \\
    0.5 & 0.999 & 1 
    & $73.07 \pm 0.20$ & $\mathbf{20.58 \pm 0.10}$
    & $134.51 \pm 0.37$  & $\mathbf{62.28 \pm 0.07}$
    \\
    0.5 & 0.999 & 2 
    & $18.74 \pm 0.24$ & $\mathbf{17.19 \pm 0.32}$
    & $40.67 \pm 0.29$ & $\mathbf{40.54 \pm 0.20}$
    \\
    0.5 & 0.999 & 5 
    & $21.10 \pm 0.89$ & $\mathbf{18.39 \pm 0.04}$
    & $84.20 \pm 0.67$ & $\mathbf{75.72 \pm 0.19}$
    \\
\bottomrule
\end{tabular}
}
\vspace{0.1cm}
\caption{Mean FID scores (lower is better) across a range of hyperparameter settings. $(\beta_1, \beta_2)$ represents the hyperparameters of the Adam optimizer, and $n_{\text{dis}}$ represents the number of discriminator steps per generator step. Our method performs robustly in a wide range of training settings \textit{without any tuning}.}
\label{stability_table}
\end{table*}

\paragraph{Improved training stability} Similar to \cite{chen2019self}, we test training stability through evaluating the sensitivity of model performance when hyperparameters are varied across a range of popular settings for training GANs, such as the Adam parameters $(\beta_1, \beta_2)$ and number of discriminator steps per generator step, $n_{\text{dis}}$, all chosen from well-tested settings in seminal GAN works \cite{chen2019self, miyato2018spectral, gulrajani2017improved, warde2016improving, radford2015unsupervised}. As seen in Table \ref{stability_table}, in comparison to SNGAN at the \textit{same architectural capacity}, InfoMax-GAN consistently improves FID for different datasets even in instances where GAN training does not converge (e.g. when $n_{\text{dis}}=1$). The FID score variability for InfoMax-GAN is much lower than SNGAN, showing its robustness to changes in training hyperparameters. Finally, while different sets of $(\beta_1, \beta_2)$ work better for each dataset, our method stabilizes training and obtain significant improvements in all these settings, \textit{without any hyperparameter tuning}. This can be useful in practice when training new GANs or on novel datasets, where training can be highly unstable when other hyperparameters are not well-tuned.

\begin{figure*}
\centering
    \begin{subfigure}{\textwidth}
        \centering
        \includegraphics[width=0.196\linewidth]{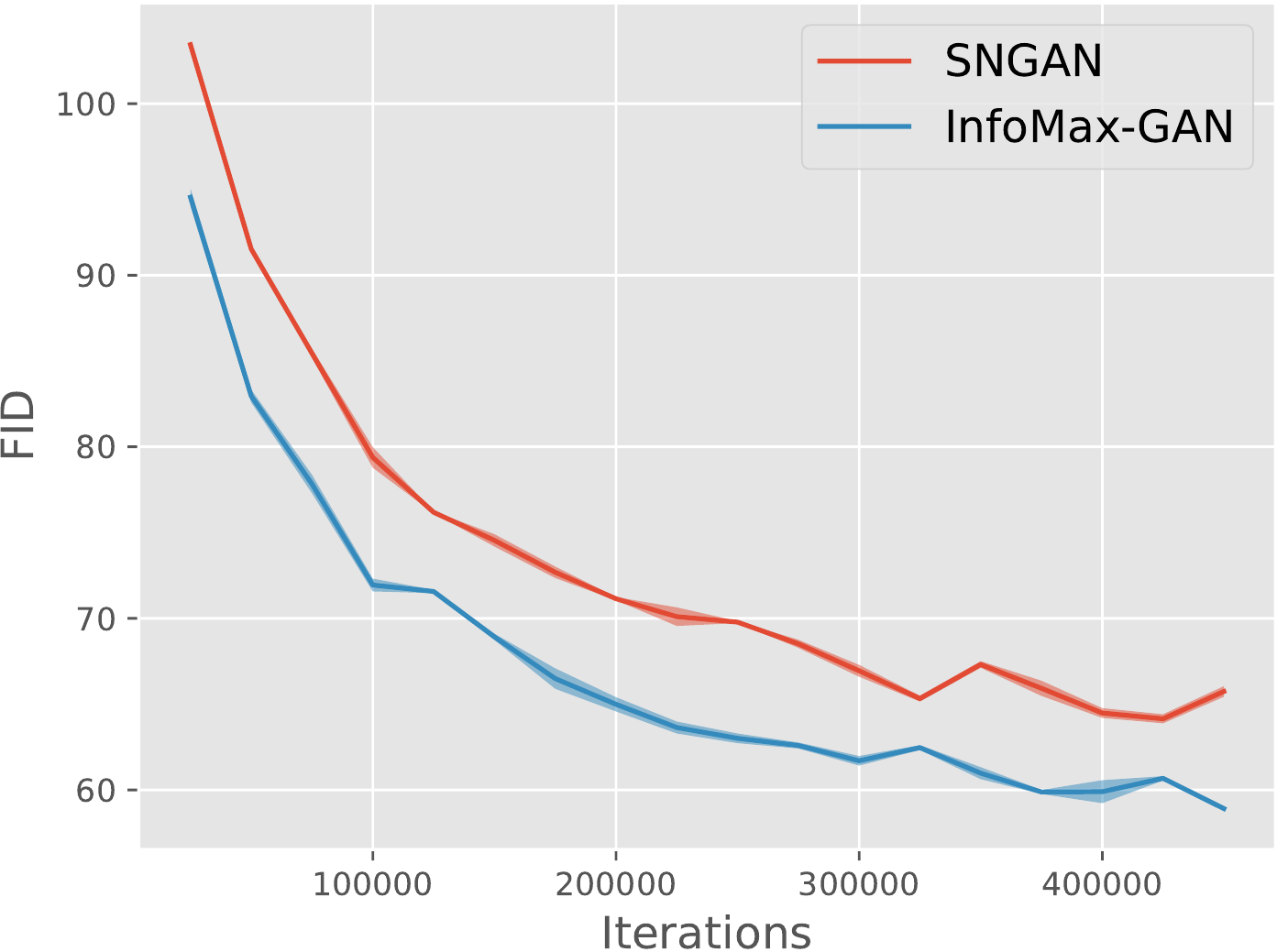}        
        \includegraphics[width=0.196\linewidth]{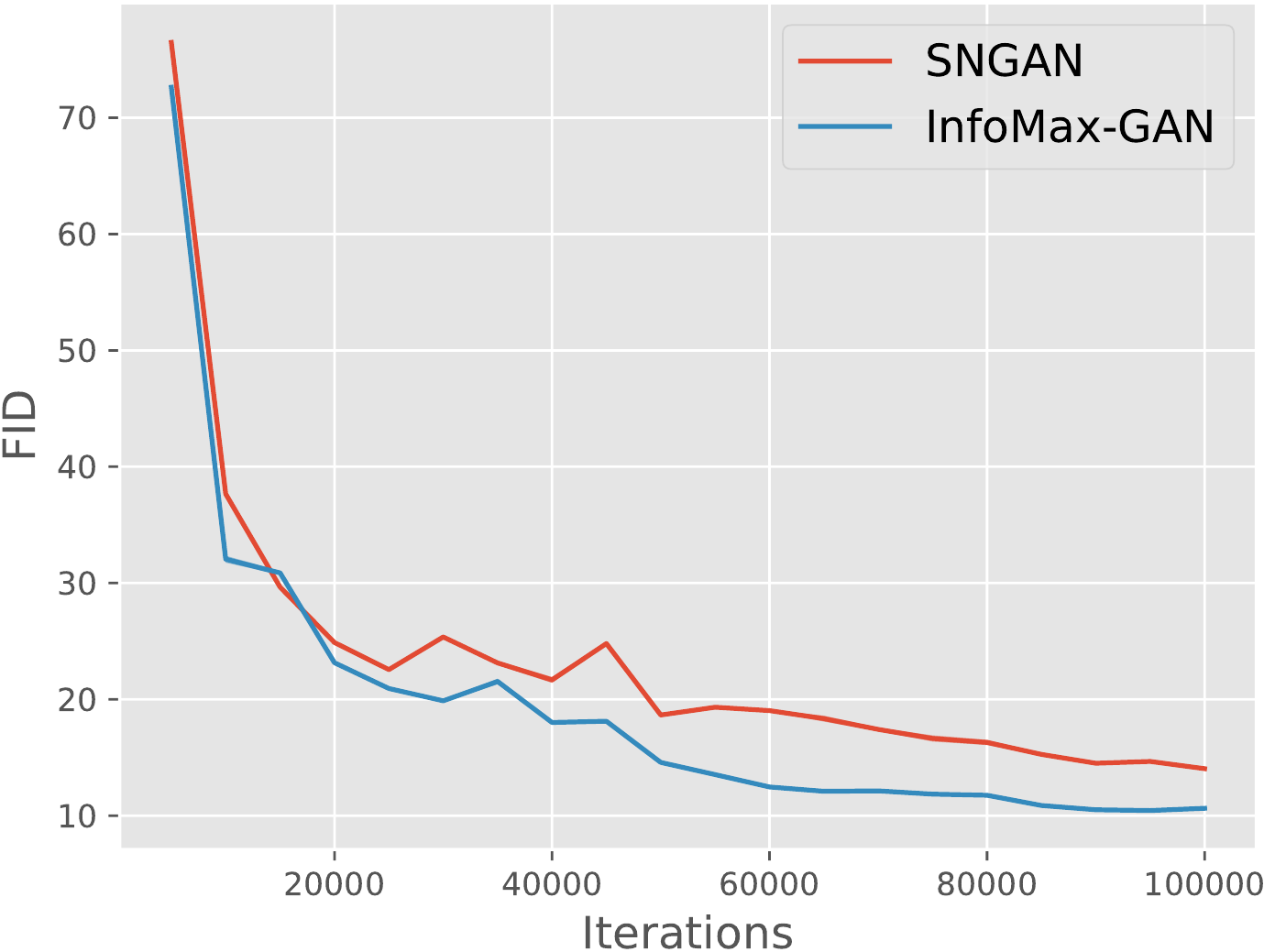}
        \includegraphics[width=0.196\linewidth]{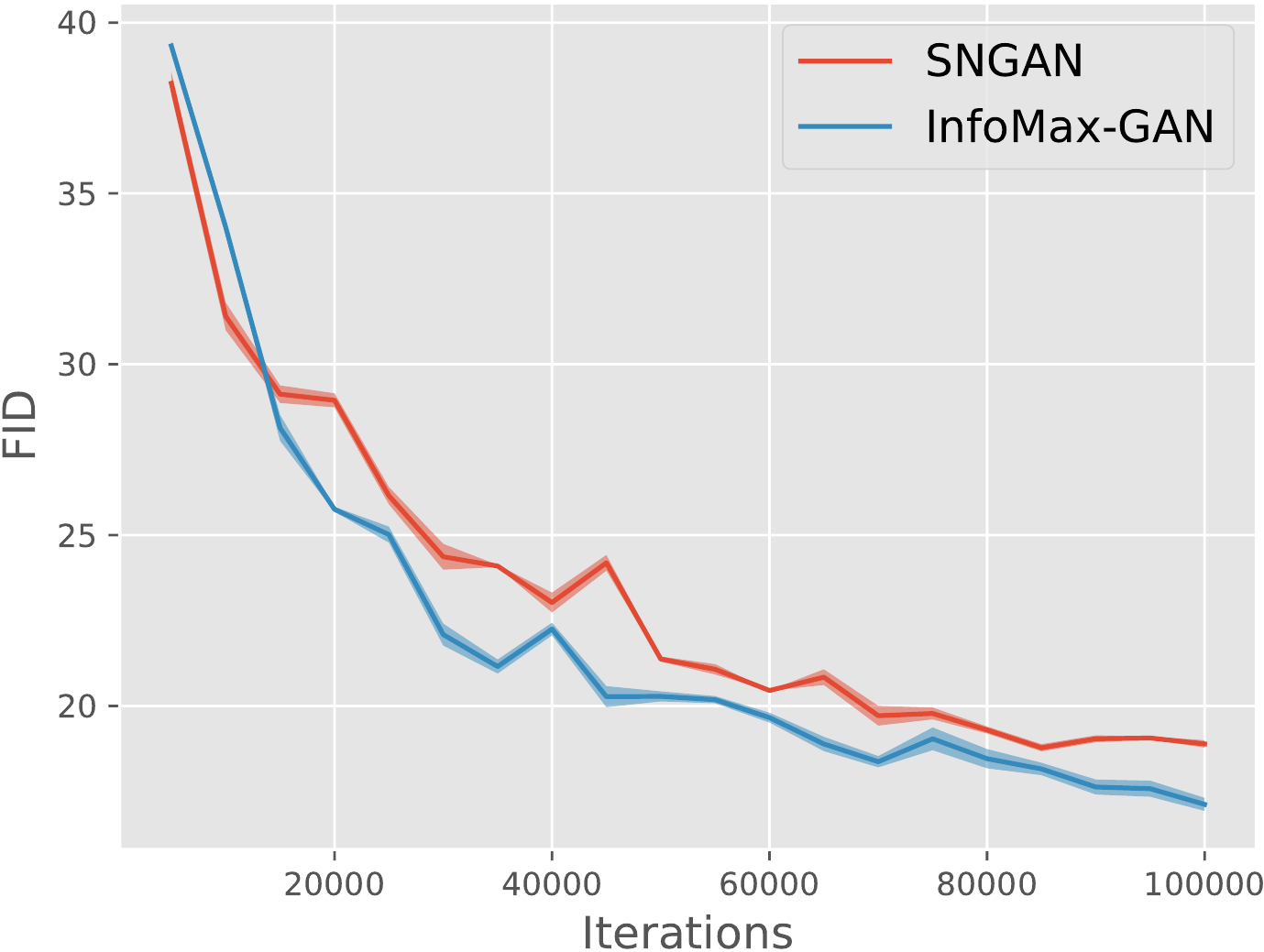}
        \includegraphics[width=0.196\linewidth]{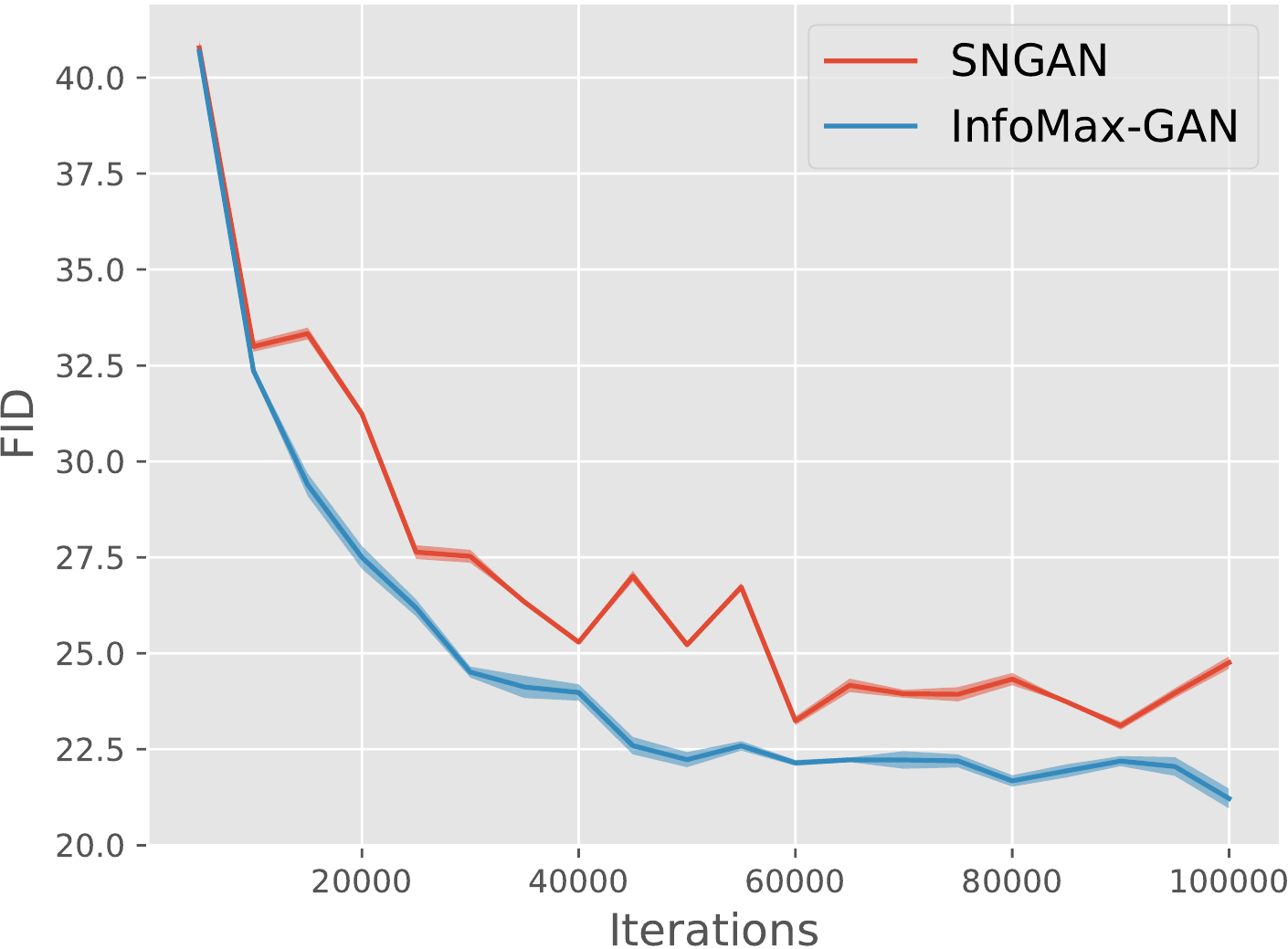}
        \includegraphics[width=0.196\linewidth]{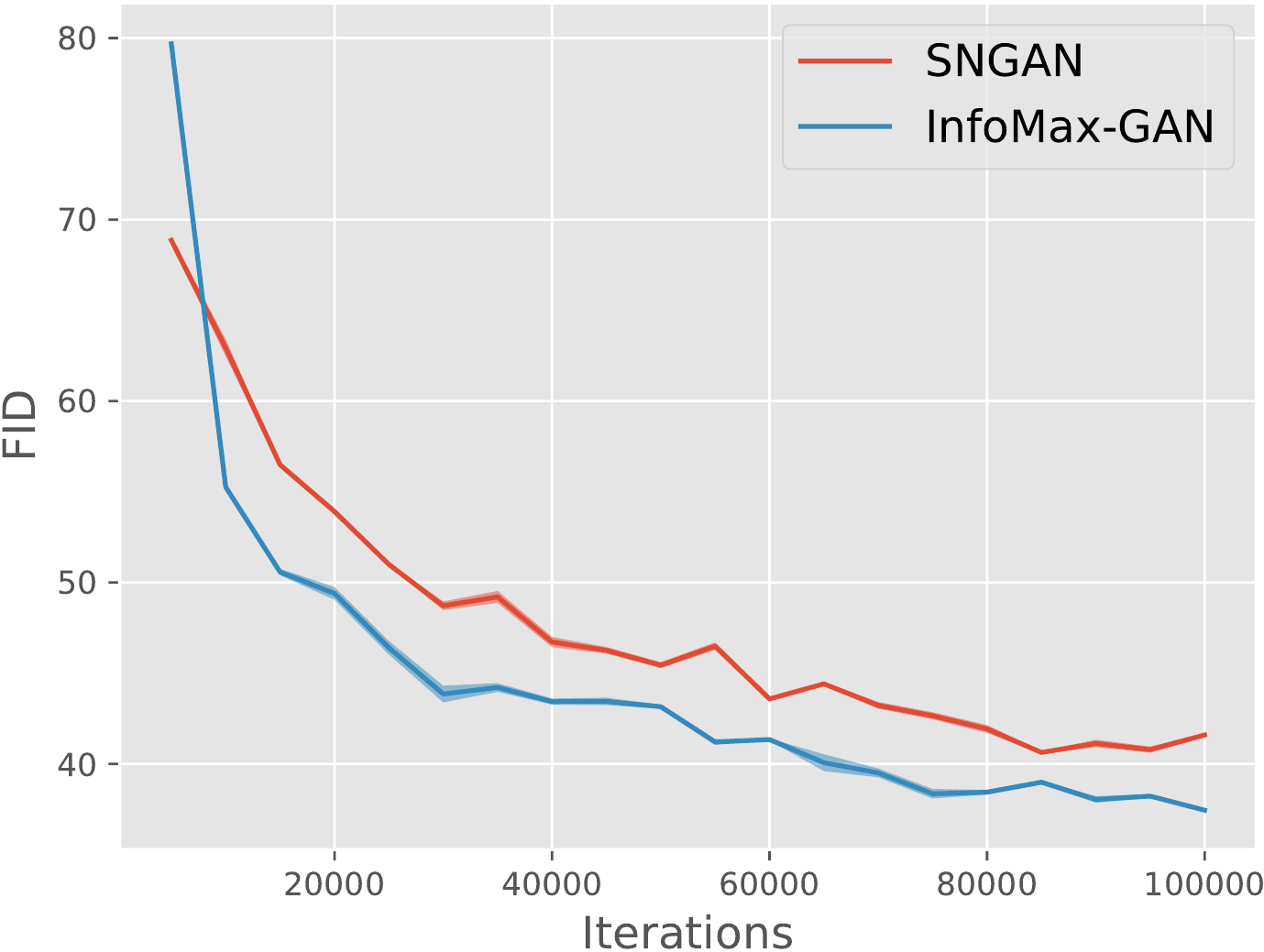}
    \end{subfigure}    
  \caption{Our approach stabilizes GAN training significantly, converges faster and consistently improves FID for all models across all datasets. Left to right: ImageNet, CelebA, CIFAR-10, CIFAR-100, STL-10.} 
  \label{fid_convergence}
\end{figure*}

In Figure \ref{fid_convergence}, we show our method stabilizes GAN training by allowing GAN training to converge faster and consistently improve performance throughout training. We attribute this to an additional constraint where the global features are constrained to have high mutual information with all their local features \cite{hjelm2018learning}, thereby constraining the space of generated data distribution and causing $p_g$ to change less radically and ultimately stabilizing the GAN training environment. This is a practical benefit when training GANs given a fixed computational budget, since significant improvements can be gained early during training.

\begin{figure}
\centering
    \begin{subfigure}{0.23\textwidth}
        \centering
        \includegraphics[width=\linewidth]{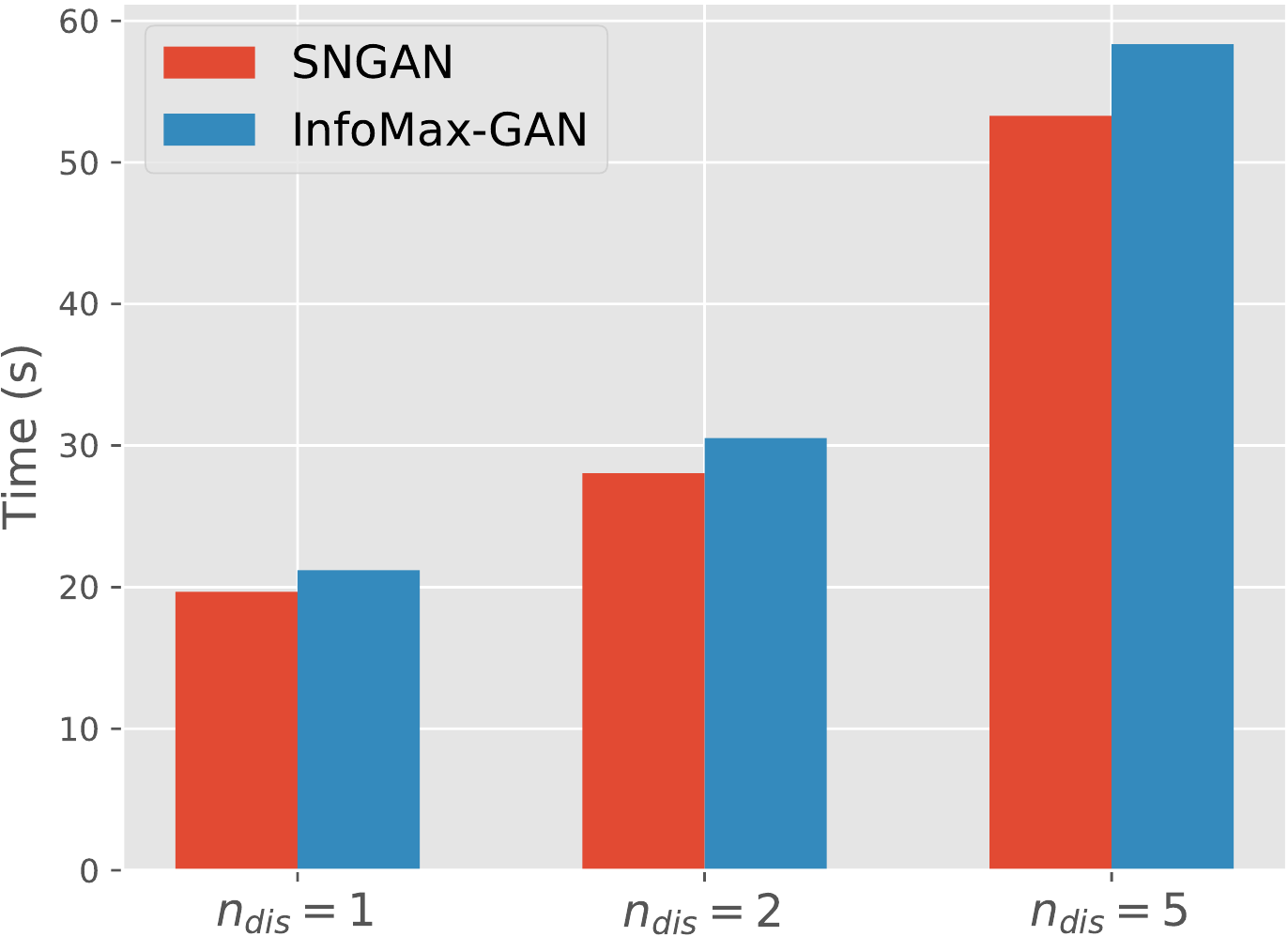}
        \caption{CIFAR-10}
    \end{subfigure}
    \begin{subfigure}{0.23\textwidth}
        \centering
        \includegraphics[width=\linewidth]{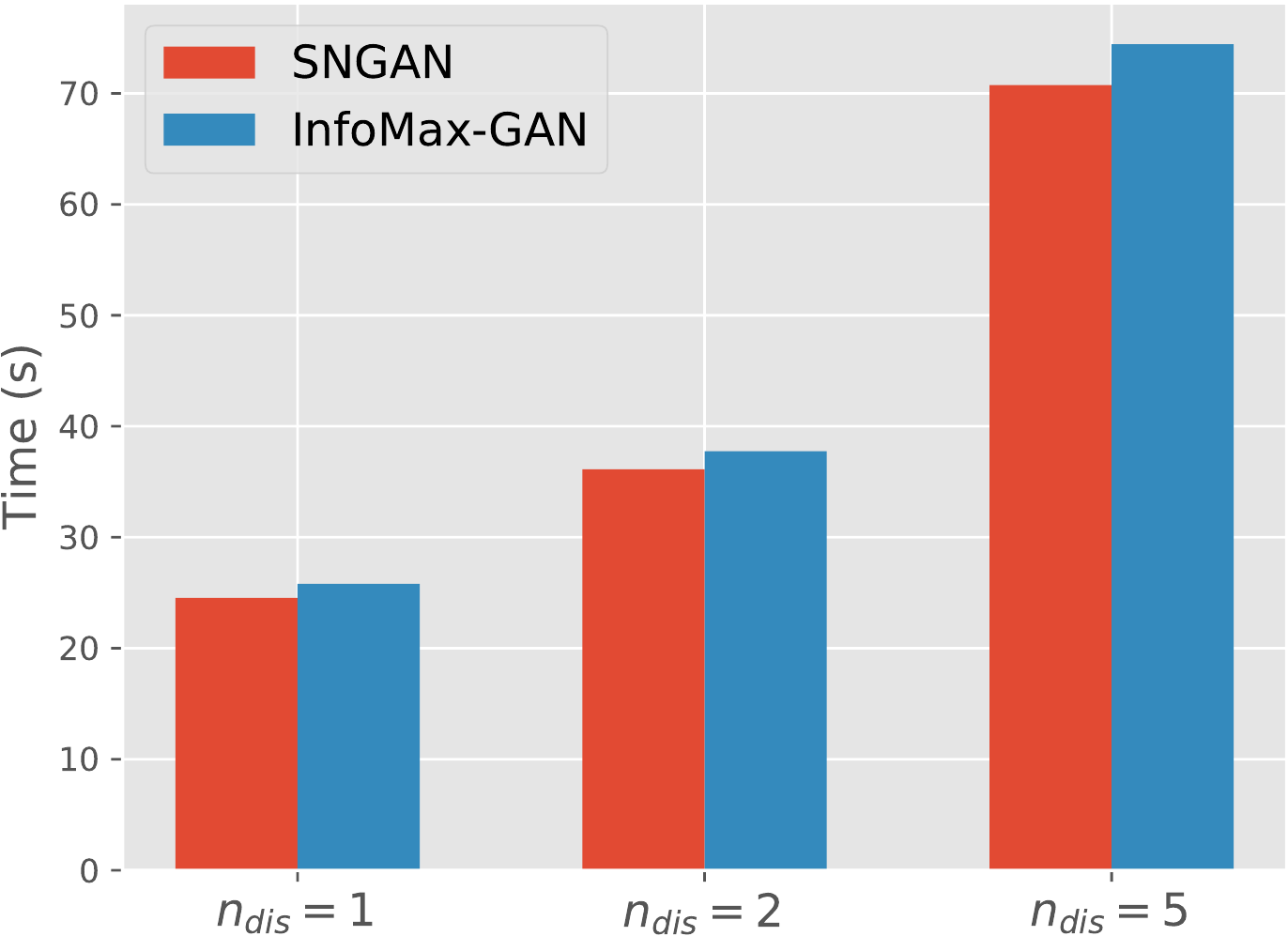}
        \caption{STL-10}
    \end{subfigure}
  \caption{Training time for $100$ generator update steps across different $n_{\text{dis}}$ values for CIFAR-10 and STL-10, using the same hardware. In general, our proposed framework incurs significantly less time than the overall training cost.}
  \label{timing_charts}
  \vspace{-0.3cm}
\end{figure}

\paragraph{Low computational cost} In practice, our method takes only a fraction of the training time. Similar to \cite{miyato2018spectral}, we profile the training time for $100$ generator update steps. In Figure \ref{timing_charts}, we see our approach takes minimal time at less than $0.1\%$ of training time per update, across all $n_{\text{dis}}$ for both CIFAR-10 and STL-10. This is since in practice, only 2 shallow (1 hidden layer) MLP networks are needed to compute the contrastive loss.  Furthermore, from Table \ref{stability_table}, at $n_{\text{dis}}=2$, InfoMax-GAN has a consistently better FID than SNGAN at $n_{\text{dis}}=5$ at approximately \textit{half the training time}, since a large $n_{\text{dis}}$ is a significant bottleneck in training time. Thus, our approach is practical for training GANs at a fixed computational budget, and has minimal computational overhead.

\paragraph{Improved mode recovery}
In Appendix \ref{appendix: mode recovery}, we demonstrate our approach helps to significantly recover more modes in the Stacked MNIST dataset \cite{metz2016unrolled}. 

\paragraph{Qualitative comparisons}
In Appendix \ref{appendix B: generated_images}, we show generated images with improved image quality against those from other models for all datasets.

\subsection{Ablation Studies}
\begin{table}
\centering
\scalebox{0.95}{
\begin{tabular}{ccc}
\toprule
	{\textbf{$R$}} &
	{\textbf{Relative Size}} &
    {\textbf{FID Score}} \\
\midrule
     256 & $2$ & $\mathbf{17.07 \pm 0.25}$ \\
     512 & $4$ & $17.21 \pm 0.15$\\
     1024 & $8$ & $17.14 \pm 0.20$ \\
     2048 & $16$ & $17.80 \pm 0.05$ \\
     4096 & $32$ &$17.38 \pm 0.11$ \\
\bottomrule
\end{tabular}
}
\vspace{0.1cm}
\caption{Mean FID scores (lower is better) for InfoMax-GAN on CIFAR-10 when the RKHS dimension $R$ is varied. Relative size here refers to how much larger $R$ is relative to the discriminator feature map depth of $128$, in terms of multiplicative factor.}
\label{rkhs_table}
\vspace{-0.3cm}
\end{table}
\paragraph{RKHS dimensions}
As seen in Table \ref{rkhs_table}, our proposed framework is robust to the choice of $R$, with the FID remaining consistent in their range of values. We attribute this to how the InfoMax critics are simple MLP networks with only 1 hidden layer, which is sufficient for achieving good representations in practice \cite{tschannen2019mutual}. We note for all our experiments in Tables \ref{main_fid}, \ref{tab:ssgan_ortho}, and \ref{stability_table}, we used $R=1024$.

\begin{figure}
\centering
\begin{subfigure}{0.23\textwidth}
  \centering
    \includegraphics[width=\linewidth]{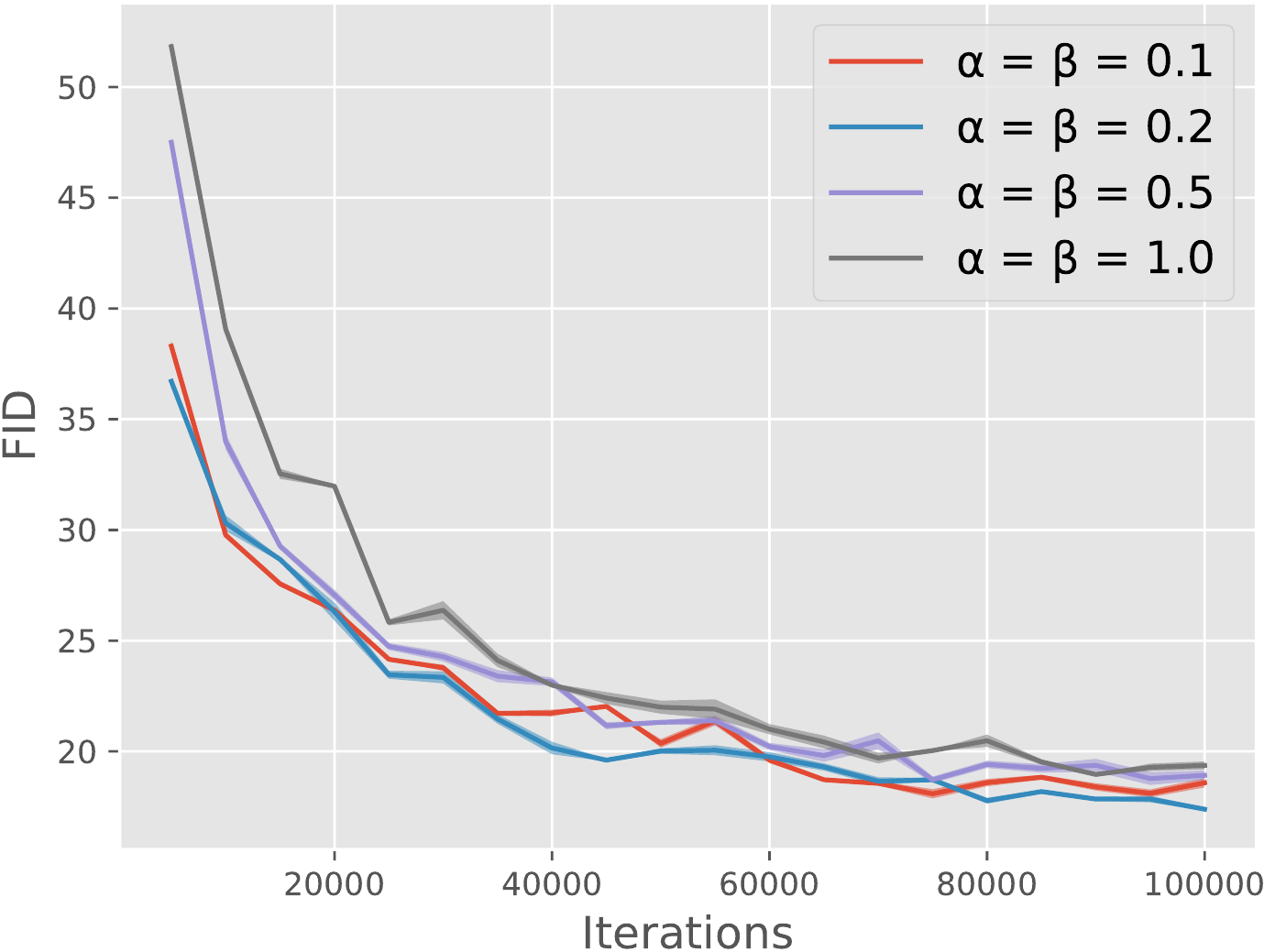}
    \caption{}
  \label{infomax_sweep:large}
\end{subfigure}
\begin{subfigure}{0.23\textwidth}
  \centering
    \includegraphics[width=\linewidth]{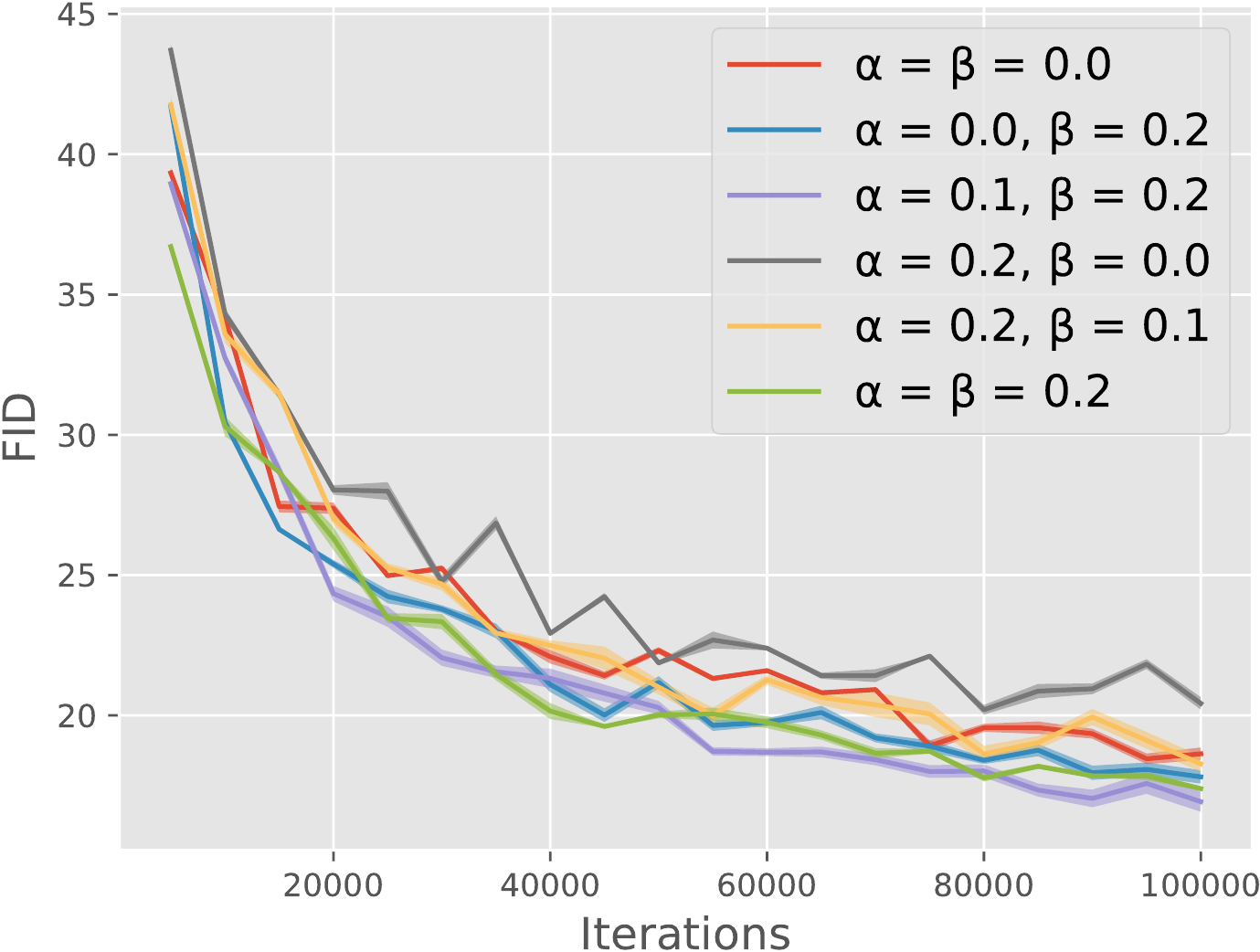}
    \caption{}
  \label{infomax_sweep:small}
\end{subfigure}
\caption{(a) CIFAR-10 FID curves for InfoMax-GAN across a large sweep of $\alpha$ and $\beta$ hyperparameters, showing $\alpha=\beta=0.2$ performs the best. (b) We perform a small sweep around the chosen hyperparameters $\alpha = \beta = 0.2$.}
\vspace{-0.5cm}
\end{figure}

\paragraph{Sensitivity of $\alpha$ and $\beta$ hyperparameters}
In Figure \ref{infomax_sweep:large}, we performed a large sweep of $\alpha$ and $\beta$ from $0.0$ to $1.0$, and see $\alpha=\beta=0.2$ obtains the best performance for our method. From Figure \ref{infomax_sweep:small}, we see our InfoMax objective for the discriminator is important for improving GAN performance: as $\beta$ is decreased, keeping $\alpha = 0.2$, FID deteriorates. Interestingly, when $\alpha=0$ and $\beta = 0.2$, having the InfoMax objective for the discriminator alone is sufficient in gaining FID improvements. This confirms our intuition of the role of information maximization in mitigating discriminator catastrophic forgetting to stabilize the GAN training environment and improve FID. However, the performance improves when the generator is also trained on the InfoMax objective, at $\alpha \in \{0.1, 0.2\}$ and $\beta=0.2$, which affirms our prior intuition that the contrastive nature of the objective helps the generator reduce mode collapse and improve FID. We note apart from this ablation study, we used $\alpha=\beta=0.2$ for all experiments reported in this paper.

\paragraph{Further studies} We include three further ablation studies on our design choices in Appendix \ref{appendix: ablation} to demonstrate the strength of our approach and justify our design choices.

\section{Related Work}
\paragraph{Mode collapse and catastrophic forgetting} Early works in reducing mode collapse include Unrolled GAN \cite{metz2016unrolled}, which restructures the generator objective with respect to unrolled discriminator optimization updates. These works often focused on assessing the number of modes recovered by a GAN based on synthetic datasets \cite{srivastava2017veegan, metz2016unrolled, che2016mode}. Subsequent works include MSGAN \cite{mao2019mode}, which introduces a regularization encouraging conditional GANs to seek out minor modes often missed when training. These works instead focus on direct metrics \cite{heusel2017gans, binkowski2018demystifying, sajjadi2018assessing, salimans2016improved, kynkaanniemi2019improved, grnarova2019domain} for assessing the diversity and quality of generated images. In our work, we utilized both types of metrics for assessment. Previous approaches to mitigate catastrophic forgetting in GANs include using forms of memory \cite{grnarova2017online, kim2018memorization, salimans2016improved}, such as checkpoint averaging. \cite{chen2019self} demonstrates the mitigation of catastrophic forgetting using a representation learning approach, which we built upon. 

\paragraph{Representation learning and GANs} To the best of our knowledge, the closest work in \textit{methodology} to ours is the state-of-the-art SSGAN, which demonstrates the use of a representation learning approach of predicting rotations \cite{gidaris2018unsupervised} to mitigate GAN forgetting and hence improve GAN performance. In contrast to SSGAN, our work uses a contrastive learning and information maximization task instead, which we demonstrate to simultaneously mitigate \textit{both} GAN forgetting and mode collapse. Furthermore, our work overcomes failure modes demonstrated in SSGAN, such as in datasets involving faces \cite{chen2019self}. For fair and accurate comparisons, our work compared with SSGAN using the \textit{exact} same architectural capacity, training and evaluation settings.

\paragraph{Information theory and GANs}
The most prominent work in utilizing mutual information maximization for GANs is InfoGAN, but we emphasize here that our work has a \textit{different focus}: while InfoGAN focuses on learning disentangled representations, our goal is to improve image synthesis. For clarity, we illustrate the specific differences with InfoGAN in Appendix \ref{appendix: infogan_comparison}. Other approaches employing information-theoretic principles include Variational GAN (VGAN) \cite{peng2018variational}, which uses an information bottleneck \cite{tishby2000information} to regularize the discriminator representations; with  \cite{bhatia2020r, mao2017least, nowozin2016f} extending to minimise divergences apart from the original JS divergence. In contrast to these works, our work employs the InfoMax principle to improve discriminator learning and provides a clear connection to how this improves GAN training via the mitigation of catastrophic forgetting.
\section{Conclusion and Future Work}
In this paper, we presented the InfoMax-GAN framework for improving natural image synthesis through simultaneously alleviating two key issues in GANs: catastrophic forgetting of the discriminator (via information maximization), and mode collapse of the generator (via contrastive learning). Our approach significantly improves on the natural image synthesis task for \textit{five} widely used datasets, and further overcome failure modes in state-of-the-art models like SSGAN. Our approach is simple and practical: it has only one auxiliary objective, performs robustly in a wide range of training settings without any hyperparameter tuning, has a low computational cost, and demonstrated improvements even when integrated to existing state-of-the-art models like SSGAN. As future work, it would be interesting to explore this framework for different tasks, such as in 3D view synthesis, where one could formulate objectives involving mutual information and adjacent views. To the best of our knowledge, our work is the first to investigate using information maximization and contrastive learning to improve GAN image synthesis performance, and we hope our work opens up new possibilities in this direction.

{\small
\bibliographystyle{ieee_fullname}
\bibliography{ms}

\begin{thebibliography}{10}\itemsep=-1pt

\bibitem{pfnet}
{G}{A}{N}s with {S}pectral {N}ormalization and {P}rojection {D}iscriminator.
\newblock \url{https://github.com/pfnet-research/sngan\_projection}.
\newblock Accessed: 2019-10-31.

\bibitem{aronszajn1950theory}
Nachman Aronszajn.
\newblock {T}heory of {R}eproducing {K}ernels.
\newblock {\em {T}ransactions of the {A}merican {M}athematical {S}ociety},
  68(3):337--404, 1950.

\bibitem{bachman2019learning}
Philip Bachman, R~Devon Hjelm, and William Buchwalter.
\newblock {L}earning {R}epresentations by {M}aximizing {M}utual {I}nformation
  {A}cross {V}iews.
\newblock {\em ar{X}iv {P}reprint ar{X}iv:1906.00910}, 2019.

\bibitem{barber2003algorithm}
David Barber and Felix~V Agakov.
\newblock {T}he {I}{M} {A}lgorithm: {A} {V}ariational {A}pproach to
  {I}nformation {M}aximization.
\newblock In {\em {A}dvances in {N}eural {I}nformation {P}rocessing {S}ystems},
  page None, 2003.

\bibitem{bengio2013representation}
Yoshua Bengio, Aaron Courville, and Pascal Vincent.
\newblock {R}epresentation {L}earning: {A} {R}eview and {N}ew {P}erspectives.
\newblock {\em {I}{E}{E}{E} {T}ransactions on {P}attern {A}nalysis and
  {M}achine {I}ntelligence}, 35(8):1798--1828, 2013.

\bibitem{bhatia2020r}
Himesh Bhatia, William Paul, Fady Alajaji, Bahman Gharesifard, and Philippe
  Burlina.
\newblock {R}$\backslash$'$\{$e$\}$ nyi {G}enerative {A}dversarial {N}etworks.
\newblock {\em arXiv preprint arXiv:2006.02479}, 2020.

\bibitem{binkowski2018demystifying}
Miko{\l}aj Bi{\'n}kowski, Dougal~J Sutherland, Michael Arbel, and Arthur
  Gretton.
\newblock {D}emystifying {M}md {G}ans.
\newblock {\em ar{X}iv {P}reprint ar{X}iv:1801.01401}, 2018.

\bibitem{che2016mode}
Tong Che, Yanran Li, Athul~Paul Jacob, Yoshua Bengio, and Wenjie Li.
\newblock {M}ode {R}egularized {G}enerative {A}dversarial {N}etworks.
\newblock {\em ar{X}iv {P}reprint ar{X}iv:1612.02136}, 2016.

\bibitem{chen2020simple}
Ting Chen, Simon Kornblith, Mohammad Norouzi, and Geoffrey Hinton.
\newblock {A} {S}imple {F}ramework for {C}ontrastive {L}earning of {V}isual
  {R}epresentations.
\newblock {\em ar{X}iv {P}reprint ar{X}iv:2002.05709}, 2020.

\bibitem{chen2019self}
Ting Chen, Xiaohua Zhai, Marvin Ritter, Mario Lucic, and Neil Houlsby.
\newblock {S}elf-{S}upervised {G}ans via {A}uxiliary {R}otation {L}oss.
\newblock In {\em {P}roceedings of the {I}{E}{E}{E} {C}onference on {C}omputer
  {V}ision and {P}attern {R}ecognition}, pages 12154--12163, 2019.

\bibitem{chen2016infogan}
Xi Chen, Yan Duan, Rein Houthooft, John Schulman, Ilya Sutskever, and Pieter
  Abbeel.
\newblock {I}nfogan: {I}nterpretable {R}epresentation {L}earning by
  {I}nformation {M}aximizing {G}enerative {A}dversarial {N}ets.
\newblock In {\em {A}dvances in {N}eural {I}nformation {P}rocessing {S}ystems},
  pages 2172--2180, 2016.

\bibitem{coates2011analysis}
Adam Coates, Andrew Ng, and Honglak Lee.
\newblock {A}n {A}nalysis of {S}ingle-{L}ayer {N}etworks in {U}nsupervised
  {F}eature {L}earning.
\newblock In {\em {P}roceedings of the {F}ourteenth {I}nternational
  {C}onference on {A}rtificial {I}ntelligence and {S}tatistics}, pages
  215--223, 2011.

\bibitem{deng2009imagenet}
Jia Deng, Wei Dong, Richard Socher, Li-Jia Li, Kai Li, and Li Fei-Fei.
\newblock {I}magenet: {A} {L}arge-{S}cale {H}ierarchical {I}mage {D}atabase.
\newblock In {\em 2009 {I}{E}{E}{E} {C}onference on {C}omputer {V}ision and
  {P}attern {R}ecognition}, pages 248--255. Ieee, 2009.

\bibitem{gidaris2018unsupervised}
Spyros Gidaris, Praveer Singh, and Nikos Komodakis.
\newblock {U}nsupervised {R}epresentation {L}earning by {P}redicting {I}mage
  {R}otations.
\newblock {\em ar{X}iv {P}reprint ar{X}iv:1803.07728}, 2018.

\bibitem{goodfellow2016nips}
Ian Goodfellow.
\newblock {N}{I}{P}{S} 2016 {T}utorial: {G}enerative {A}dversarial {N}etworks.
\newblock {\em ar{X}iv {P}reprint ar{X}iv:1701.00160}, 2016.

\bibitem{goodfellow2014generative}
Ian Goodfellow, Jean Pouget-Abadie, Mehdi Mirza, Bing Xu, David Warde-Farley,
  Sherjil Ozair, Aaron Courville, and Yoshua Bengio.
\newblock {G}enerative {A}dversarial {N}ets.
\newblock In {\em {A}dvances in {N}eural {I}nformation {P}rocessing {S}ystems},
  pages 2672--2680, 2014.

\bibitem{gretton2012kernel}
Arthur Gretton, Karsten~M Borgwardt, Malte~J Rasch, Bernhard Sch{\"o}lkopf, and
  Alexander Smola.
\newblock {A} {K}ernel {T}wo-{S}ample {T}est.
\newblock {\em {J}ournal of {M}achine {L}earning {R}esearch}, 13(Mar):723--773,
  2012.

\bibitem{grnarova2017online}
Paulina Grnarova, Kfir~Y Levy, Aurelien Lucchi, Thomas Hofmann, and Andreas
  Krause.
\newblock {A}n {O}nline {L}earning {A}pproach to {G}enerative {A}dversarial
  {N}etworks.
\newblock {\em ar{X}iv {P}reprint ar{X}iv:1706.03269}, 2017.

\bibitem{grnarova2019domain}
Paulina Grnarova, Kfir~Y Levy, Aurelien Lucchi, Nathanael Perraudin, Ian
  Goodfellow, Thomas Hofmann, and Andreas Krause.
\newblock {A} {D}omain {A}gnostic {M}easure for {M}onitoring and {E}valuating
  {G}{A}{N}s.
\newblock In {\em {A}dvances in {N}eural {I}nformation {P}rocessing {S}ystems},
  pages 12069--12079, 2019.

\bibitem{gulrajani2017improved}
Ishaan Gulrajani, Faruk Ahmed, Martin Arjovsky, Vincent Dumoulin, and Aaron~C
  Courville.
\newblock {I}mproved {T}raining of {W}asserstein {G}ans.
\newblock In {\em {A}dvances in {N}eural {I}nformation {P}rocessing {S}ystems},
  pages 5767--5777, 2017.

\bibitem{he2016deep}
Kaiming He, Xiangyu Zhang, Shaoqing Ren, and Jian Sun.
\newblock {D}eep {R}esidual {L}earning for {I}mage {R}ecognition.
\newblock In {\em {P}roceedings of the {I}{E}{E}{E} {C}onference on {C}omputer
  {V}ision and {P}attern {R}ecognition}, pages 770--778, 2016.

\bibitem{henaff2019data}
Olivier~J H{\'e}naff, Ali Razavi, Carl Doersch, SM Eslami, and Aaron van~den
  Oord.
\newblock {D}ata-{E}fficient {I}mage {R}ecognition with {C}ontrastive
  {P}redictive {C}oding.
\newblock {\em ar{X}iv {P}reprint ar{X}iv:1905.09272}, 2019.

\bibitem{heusel2017gans}
Martin Heusel, Hubert Ramsauer, Thomas Unterthiner, Bernhard Nessler, and Sepp
  Hochreiter.
\newblock {G}ans {T}rained by a {T}wo {T}ime-{S}cale {U}pdate {R}ule {C}onverge
  to a {L}ocal {N}ash {E}quilibrium.
\newblock In {\em {A}dvances in {N}eural {I}nformation {P}rocessing {S}ystems},
  pages 6626--6637, 2017.

\bibitem{hjelm2018learning}
R~Devon Hjelm, Alex Fedorov, Samuel Lavoie-Marchildon, Karan Grewal, Phil
  Bachman, Adam Trischler, and Yoshua Bengio.
\newblock {L}earning {D}eep {R}epresentations by {M}utual {I}nformation
  {E}stimation and {M}aximization.
\newblock {\em ar{X}iv {P}reprint ar{X}iv:1808.06670}, 2018.

\bibitem{kemker2018measuring}
Ronald Kemker, Marc McClure, Angelina Abitino, Tyler~L Hayes, and Christopher
  Kanan.
\newblock {M}easuring {C}atastrophic {F}orgetting in {N}eural {N}etworks.
\newblock In {\em {T}hirty-{S}econd {A}{A}{A}{I} {C}onference on {A}rtificial
  {I}ntelligence}, 2018.

\bibitem{kim2018memorization}
Youngjin Kim, Minjung Kim, and Gunhee Kim.
\newblock {M}emorization {P}recedes {G}eneration: {L}earning {U}nsupervised
  {G}ans with {M}emory {N}etworks.
\newblock {\em ar{X}iv {P}reprint ar{X}iv:1803.01500}, 2018.

\bibitem{kingma2014adam}
Diederik~P Kingma and Jimmy Ba.
\newblock {A}dam: {A} {M}ethod for {S}tochastic {O}ptimization.
\newblock {\em ar{X}iv {P}reprint ar{X}iv:1412.6980}, 2014.

\bibitem{kirkpatrick2017overcoming}
James Kirkpatrick, Razvan Pascanu, Neil Rabinowitz, Joel Veness, Guillaume
  Desjardins, Andrei~A Rusu, Kieran Milan, John Quan, Tiago Ramalho, Agnieszka
  Grabska-Barwinska, et~al.
\newblock {O}vercoming {C}atastrophic {F}orgetting in {N}eural {N}etworks.
\newblock {\em {P}roceedings of the {N}ational {A}cademy of {S}ciences},
  114(13):3521--3526, 2017.

\bibitem{kong2019mutual}
Lingpeng Kong, Cyprien de~Masson d'Autume, Wang Ling, Lei Yu, Zihang Dai, and
  Dani Yogatama.
\newblock {A} {M}utual {I}nformation {M}aximization {P}erspective of {L}anguage
  {R}epresentation {L}earning.
\newblock {\em ar{X}iv {P}reprint ar{X}iv:1910.08350}, 2019.

\bibitem{krizhevsky2009learning}
Alex Krizhevsky, Geoffrey Hinton, et~al.
\newblock {L}earning {M}ultiple {L}ayers of {F}eatures from {T}iny {I}mages.
\newblock Technical report, Citeseer, 2009.

\bibitem{kurach2019large}
{K}arol {K}urach, {M}ario {L}u{\v{c}}i{\'c}, {X}iaohua {Z}hai, {M}arcin
  {M}ichalski, and {S}ylvain {G}elly.
\newblock {A} {L}arge-scale {S}tudy on {R}egularization and {N}ormalization in
  {G}{A}{N}s.
\newblock In {\em {I}nternational {C}onference on {M}achine {L}earning}, pages
  3581--3590, 2019.

\bibitem{kynkaanniemi2019improved}
Tuomas Kynk{\"a}{\"a}nniemi, Tero Karras, Samuli Laine, Jaakko Lehtinen, and
  Timo Aila.
\newblock {I}mproved {P}recision and {R}ecall {M}etric for {A}ssessing
  {G}enerative {M}odels.
\newblock In {\em {A}dvances in {N}eural {I}nformation {P}rocessing {S}ystems},
  pages 3929--3938, 2019.

\bibitem{lecun1998mnist}
Yann LeCun.
\newblock {T}he {M}{N}{I}{S}{T} {D}atabase of {H}andwritten {D}igits.
\newblock {\em {H}ttp://yann. {L}ecun. {C}om/{E}xdb/{M}nist/}, 1998.

\bibitem{lee2020mimicry}
Kwot~Sin Lee and Christopher Town.
\newblock {M}imicry: {T}owards the {R}eproducibility of {G}{A}{N} {R}esearch.
\newblock {\em {C}{V}{P}{R} AI for Content Creation Workshop}, 2020.

\bibitem{linsker1988self}
Ralph Linsker.
\newblock {S}elf-{O}rganization in a {P}erceptual {N}etwork.
\newblock {\em {C}omputer}, 21(3):105--117, 1988.

\bibitem{liu2015faceattributes}
Ziwei Liu, Ping Luo, Xiaogang Wang, and Xiaoou Tang.
\newblock {D}eep {L}earning {F}ace {A}ttributes in the {W}ild.
\newblock In {\em {P}roceedings of {I}nternational {C}onference on {C}omputer
  {V}ision ({I}{C}{C}{V})}, December 2015.

\bibitem{lowe2019greedy}
Sindy L{\"o}we, Peter O'Connor, and Bastiaan~S Veeling.
\newblock {G}reedy {I}nfo{M}ax for {B}iologically {P}lausible
  {S}elf-{S}upervised {R}epresentation {L}earning.
\newblock {\em ar{X}iv {P}reprint ar{X}iv:1905.11786}, 2019.

\bibitem{mao2019mode}
Qi Mao, Hsin-Ying Lee, Hung-Yu Tseng, Siwei Ma, and Ming-Hsuan Yang.
\newblock {M}ode {S}eeking {G}enerative {A}dversarial {N}etworks for {D}iverse
  {I}mage {S}ynthesis.
\newblock In {\em {P}roceedings of the {I}{E}{E}{E} {C}onference on {C}omputer
  {V}ision and {P}attern {R}ecognition}, pages 1429--1437, 2019.

\bibitem{mao2017least}
Xudong Mao, Qing Li, Haoran Xie, Raymond~Y.K. Lau, Zhen Wang, and Stephen
  Paul~Smolley.
\newblock {L}east {S}quares {G}enerative {A}dversarial {N}etworks.
\newblock In {\em {T}he {I}{E}{E}{E} {I}nternational {C}onference on {C}omputer
  {V}ision ({I}{C}{C}{V})}, Oct 2017.

\bibitem{mccloskey1989catastrophic}
Michael McCloskey and Neal~J Cohen.
\newblock {C}atastrophic {I}nterference in {C}onnectionist {N}etworks: {T}he
  {S}equential {L}earning {P}roblem.
\newblock In {\em {P}sychology of {L}earning and {M}otivation}, volume~24,
  pages 109--165. Elsevier, 1989.

\bibitem{metz2016unrolled}
Luke Metz, Ben Poole, David Pfau, and Jascha Sohl-Dickstein.
\newblock {U}nrolled {G}enerative {A}dversarial {N}etworks.
\newblock {\em ar{X}iv {P}reprint ar{X}iv:1611.02163}, 2016.

\bibitem{miyato2018spectral}
Takeru Miyato, Toshiki Kataoka, Masanori Koyama, and Yuichi Yoshida.
\newblock {S}pectral {N}ormalization for {G}enerative {A}dversarial {N}etworks.
\newblock {\em ar{X}iv {P}reprint ar{X}iv:1802.05957}, 2018.

\bibitem{miyato2018cgans}
Takeru Miyato and Masanori Koyama.
\newblock c{G}{A}{N}s with {P}rojection {D}iscriminator.
\newblock {\em ar{X}iv {P}reprint ar{X}iv:1802.05637}, 2018.

\bibitem{nowozin2016f}
Sebastian Nowozin, Botond Cseke, and Ryota Tomioka.
\newblock f-{G}{A}{N}: {T}raining {G}enerative {N}eural {S}amplers using
  {V}ariational {D}ivergence {M}inimization.
\newblock In {\em {A}dvances in {N}eural {I}nformation {P}rocessing {S}ystems},
  pages 271--279, 2016.

\bibitem{odena2019open}
Augustus Odena.
\newblock {O}pen {Q}uestions {A}bout {G}enerative {A}dversarial {N}etworks.
\newblock {\em {D}istill}, 2019.
\newblock https://distill.pub/2019/gan-open-problems.

\bibitem{odena2017conditional}
Augustus Odena, Christopher Olah, and Jonathon Shlens.
\newblock {C}onditional {I}mage {S}ynthesis with {A}uxiliary {C}lassifier
  {G}ans.
\newblock In {\em {P}roceedings of the 34th {I}nternational {C}onference on
  {M}achine {L}earning-{V}olume 70}, pages 2642--2651. JMLR. org, 2017.

\bibitem{oord2018representation}
Aaron van~den Oord, Yazhe Li, and Oriol Vinyals.
\newblock {R}epresentation {L}earning with {C}ontrastive {P}redictive {C}oding.
\newblock {\em ar{X}iv {P}reprint ar{X}iv:1807.03748}, 2018.

\bibitem{ozair2019wasserstein}
{S}herjil {O}zair, {C}orey {L}ynch, {Y}oshua {B}engio, {A}aron {V}an~den
  {O}ord, {S}ergey {L}evine, and {P}ierre {S}ermanet.
\newblock {W}asserstein {D}ependency {M}easure for {R}epresentation {L}earning.
\newblock In {\em {A}dvances in {N}eural {I}nformation {P}rocessing {S}ystems},
  pages 15604--15614, 2019.

\bibitem{paninski2003estimation}
Liam Paninski.
\newblock {E}stimation of {E}ntropy and {M}utual {I}nformation.
\newblock {\em {N}eural {C}omputation}, 15(6):1191--1253, 2003.

\bibitem{peng2018variational}
Xue~Bin Peng, Angjoo Kanazawa, Sam Toyer, Pieter Abbeel, and Sergey Levine.
\newblock {V}ariational {D}iscriminator {B}ottleneck: {I}mproving {I}mitation
  {L}earning, {I}nverse {R}l, and {G}ans by {C}onstraining {I}nformation
  {F}low.
\newblock {\em ar{X}iv {P}reprint ar{X}iv:1810.00821}, 2018.

\bibitem{poole2018variational}
Ben Poole, Sherjil Ozair, A{\"a}ron van~den Oord, Alexander~A Alemi, and George
  Tucker.
\newblock {O}n {V}ariational {L}ower {B}ounds of {M}utual {I}nformation.
\newblock In {\em {N}eur{I}{P}{S} {W}orkshop on {B}ayesian {D}eep {L}earning},
  2018.

\bibitem{radford2015unsupervised}
Alec Radford, Luke Metz, and Soumith Chintala.
\newblock {U}nsupervised {R}epresentation {L}earning with {D}eep
  {C}onvolutional {G}enerative {A}dversarial {N}etworks.
\newblock {\em ar{X}iv {P}reprint ar{X}iv:1511.06434}, 2015.

\bibitem{sajjadi2018assessing}
Mehdi~SM Sajjadi, Olivier Bachem, Mario Lucic, Olivier Bousquet, and Sylvain
  Gelly.
\newblock {A}ssessing {G}enerative {M}odels via {P}recision and {R}ecall.
\newblock In {\em {A}dvances in {N}eural {I}nformation {P}rocessing {S}ystems},
  pages 5228--5237, 2018.

\bibitem{salimans2016improved}
Tim Salimans, Ian Goodfellow, Wojciech Zaremba, Vicki Cheung, Alec Radford, and
  Xi Chen.
\newblock {I}mproved {T}echniques for {T}raining {G}ans.
\newblock In {\em {A}dvances in {N}eural {I}nformation {P}rocessing {S}ystems},
  pages 2234--2242, 2016.

\bibitem{srivastava2017veegan}
Akash Srivastava, Lazar Valkov, Chris Russell, Michael~U Gutmann, and Charles
  Sutton.
\newblock {V}eegan: {R}educing {M}ode {C}ollapse in {G}ans {U}sing {I}mplicit
  {V}ariational {L}earning.
\newblock In {\em {A}dvances in {N}eural {I}nformation {P}rocessing {S}ystems},
  pages 3308--3318, 2017.

\bibitem{szegedy2016rethinking}
Christian Szegedy, Vincent Vanhoucke, Sergey Ioffe, Jon Shlens, and Zbigniew
  Wojna.
\newblock {R}ethinking the {I}nception {A}rchitecture for {C}omputer {V}ision.
\newblock In {\em {P}roceedings of the {I}{E}{E}{E} {C}onference on {C}omputer
  {V}ision and {P}attern {R}ecognition}, pages 2818--2826, 2016.

\bibitem{tian2019contrastive}
Yonglong Tian, Dilip Krishnan, and Phillip Isola.
\newblock {C}ontrastive {M}ultiview {C}oding.
\newblock {\em ar{X}iv {P}reprint ar{X}iv:1906.05849}, 2019.

\bibitem{tishby2000information}
Naftali Tishby, Fernando~C Pereira, and William Bialek.
\newblock {T}he {I}nformation {B}ottleneck {M}ethod.
\newblock {\em ar{X}iv {P}reprint {P}hysics/0004057}, 2000.

\bibitem{Tran_2018_ECCV}
Ngoc-Trung Tran, Tuan-Anh Bui, and Ngai-Man Cheung.
\newblock {D}ist-{G}an: {A}n {I}mproved {G}{A}{N} {U}sing {D}istance
  {C}onstraints.
\newblock In {\em {T}he {E}uropean {C}onference on {C}omputer {V}ision
  ({E}{C}{C}{V})}, September 2018.

\bibitem{tran2019improving}
Ngoc-Trung Tran, Tuan-Anh Bui, and Ngai-Man Cheung.
\newblock {I}mproving {G}{A}{N} with {N}eighbors {E}mbedding and {G}radient
  {M}atching.
\newblock In {\em {P}roceedings of the {A}{A}{A}{I} {C}onference on
  {A}rtificial {I}ntelligence}, volume~33, pages 5191--5198, 2019.

\bibitem{tran2019self}
Ngoc-Trung Tran, Viet-Hung Tran, Bao-Ngoc Nguyen, Linxiao Yang, et~al.
\newblock {S}elf-{S}upervised {G}{A}{N}: {A}nalysis and {I}mprovement with
  {M}ulti-{C}lass {M}inimax {G}ame.
\newblock In {\em {A}dvances in {N}eural {I}nformation {P}rocessing {S}ystems},
  pages 13232--13243, 2019.

\bibitem{tran2019improved}
Ngoc-Trung Tran, Viet-Hung Tran, Ngoc-Bao Nguyen, and Ngai-Man Cheung.
\newblock {A}n {I}mproved {S}elf-supervised {G}{A}{N} via {A}dversarial
  {T}raining.
\newblock {\em arXiv preprint arXiv:1905.05469}, 2019.

\bibitem{tschannen2019mutual}
Michael Tschannen, Josip Djolonga, Paul~K Rubenstein, Sylvain Gelly, and Mario
  Lucic.
\newblock {O}n {M}utual {I}nformation {M}aximization for {R}epresentation
  {L}earning.
\newblock {\em ar{X}iv {P}reprint ar{X}iv:1907.13625}, 2019.

\bibitem{warde2016improving}
David Warde-Farley and Yoshua Bengio.
\newblock {I}mproving {G}enerative {A}dversarial {N}etworks with {D}enoising
  {F}eature {M}atching.
\newblock 2016.

\end{thebibliography}
}

\clearpage
\appendix
\section{Supplementary Results}
\label{appendix:A}

\subsection{Datasets}
\label{appendix: dataset_settings}
We evaluate our models across five different datasets:
ImageNet \cite{deng2009imagenet}, CelebA \cite{liu2015faceattributes}, CIFAR-10 \cite{krizhevsky2009learning}, STL-10 \cite{coates2011analysis}, and CIFAR-100 \cite{krizhevsky2009learning}. In general, for preprocessing our images, we follow settings in \cite{miyato2018cgans, chen2019self}. Specifically, for ImageNet, we use the 1.3M training images downsampled to size $128 \times 128$. For CelebA, we use the aligned version of the 200K images downsampled to size $128 \times 128$. For CIFAR-10 and CIFAR-100 we use all 50K training images, and for STL-10, we use all 100K unlabeled images downsampled to size $48 \times 48$.

\subsection{Training Settings}
\label{appendix: training_settings}
For all models, we use Residual Network \cite{he2016deep} backbones following \cite{miyato2018spectral}. For training the models on all datasets, we adopt the Adam optimizer \cite{kingma2014adam} with a learning rate of $2\times10^{-4}$ and batch size of $64$, following \cite{gulrajani2017improved, miyato2018spectral}. Specifically, for CIFAR-10, CIFAR-100 and STL-10, we follow settings in \cite{miyato2018cgans} by linearly decaying learning rate over 100K generator steps, each taken every 5 discriminator update steps. For ImageNet, we follow \cite{miyato2018spectral} by increasing the number of generator updates to 450K steps instead, but with no learning rate decay. For CelebA, we follow \cite{chen2019self} by taking 100K generator steps, each taken after 2 discriminator updates and with no learning rate decay.

We emphasize that for \textit{fairness} in our comparisons, we re-implemented all considered models using the same code base and framework, and trained all models under the \textit{exact same training conditions} for each dataset.

\subsection{Evaluation Settings}
\label{appendix: eval_metrics}
In our work, we use three different evaluation metrics: Fréchet Inception Distance (FID) \cite{heusel2017gans} and Kernel Inception Distance (KID) \cite{binkowski2018demystifying} to evaluate generated image diversity, and Inception Score (IS) \cite{salimans2016improved} to evaluate image quality.

\paragraph{Fréchet Inception Distance} Firstly, FID is a popular metric measuring the diversity of generated images, which we adopt for ease of comparisons since it is widely used in the literature. Formally, FID computes the Wasserstein-2 Distance between features produced by a pre-trained Inception \cite{szegedy2016rethinking} network for input real and generated images, and is defined as:
\begin{equation}
    d_{\text{FID}} = \norm{\mu_r - \mu_g}^2_2 + \Tr{(\Sigma_r + \Sigma_g - 2(\Sigma_r \Sigma_g)^{\frac{1}{2}})}
\end{equation}
where $\mu_r$ and $\Sigma_r$ denotes the mean and covariance of feature vectors produced by forwarding real images through a pre-trained Inception \cite{szegedy2016rethinking} network, and $\mu_g$ and $\Sigma_g$ similarly represents the equivalent for fake images. Intuitively, FID measures the diversity of the generated images, since the features of the generated images should ideally have a small distance with those of real images if they look similar on average. However, we note that FID can produce highly biased estimates \cite{binkowski2018demystifying}, where using larger sample sizes can produce better scores, which can causes FID comparisons to be often mismatched \cite{kurach2019large} in practice. Thus, we emphasize for fairness in comparisons, we use the \textit{exact} same number of real and fake images for computing FID. 

\paragraph{Kernel Inception Distance} KID is an alternative metric highly correlated with FID that also measures diversity of images, but produces unbiased estimates \cite{binkowski2018demystifying}, which is useful for corroborating our findings on FID. Formally, KID measures the square of the Maximum Mean Discrepancy (MMD) \cite{gretton2012kernel} between two probability distributions in a metric space, and can be defined as:
\begin{equation}
    \begin{split}
        d_{\text{KID}} &= \text{MMD}^2(X, Y) \\
            &= \frac{1}{m(m-1)} \sum^m_{i \neq j}k(x_i, x_j) \\
            &+ \frac{1}{n(n-1)}\sum^n_{i \neq j}k(y_i, y_j)
            - \frac{2}{mn}\sum^m_{i=1}\sum^n_{j=1}k(x_i, y_j)
    \end{split}
\end{equation}
for two random variables $X$ and $Y$ from different distributions, sample sizes $m$ and $n$, and $k$ is the polynomial kernel defined as:
\begin{equation}
    k(x, y) = (\frac{1}{d} x^Ty + 1)^3
\end{equation}
where $d$ represents the dimensions of the samples. Intuitively, MMD measures the distance between distributions using a function from a class of witness functions such that if the true distance between the distributions is zero, the distance between the mean embeddings produced by this function will also be zero. Here, the polynomial kernel is cubic in order to measure the first three moments of the distributions (mean, variance, and skewness), and the embedding is defined on the feature space through the Inception network. Similar to FID, we use the same number of real and fake images for all models when computing KID.

\paragraph{Inception Score} Finally, IS aims to measure the realism of generated images using the same Inception network, and can be formally defined as:
\begin{equation}
    d_{IS} = \exp (\mathbb{E}_{x \sim p_g} \mathcal{D}_{\text{KL}}(p(y\vert x) \vert\vert p(y)))
\end{equation}
where a high score is achieved if the conditional class distribution $p(y|x)$ has low entropy and the marginal class distribution $p(y)$ has high entropy, causing a large KL divergence between the two distributions for some samples $x$ from the generated image distribution $p_g$. Intuitively, a large score is produced if the Inception network gives a high probability to one class, indicating it looks realistically in one class. Thus, IS tends to correlate well with human assessment for quality of images \cite{salimans2016improved}.

\paragraph{Sample sizes}
For all FID scores reported in this paper, we compute them using 50K real samples and 10K fake samples across 3 random seeds to report the mean and standard deviation of the scores. As 50K real samples are much lesser than the 1.3M images in ImageNet, we randomly sample without replacement $50$ images from each of the $1000$ classes to compute the real image statistics, to avoid high bias in the results. We emphasize that for fairness in comparisons, we used the same number of real and fake samples when computing FID, since FID can produce highly bias estimates at different sample sizes \cite{binkowski2018demystifying}. In fact, we note that \textit{lower FID scores} can indeed be obtained if we simply use larger sample sizes, particularly for larger datasets like ImageNet. However, our experiments show that in practice, the performance \textit{margins} remain the same above our current configuration. For KID, we follow the same procedure for all datasets but use 50K real and fake samples instead. Finally, for IS, we use 50K fake samples. 

We emphasize that all these evaluation settings are kept the same for all model evaluations for each dataset, in order to ensure fairness and accuracy in our comparisons.

\subsection{Improved Mode Recovery}
\label{appendix: mode recovery}
\begin{table}
\centering
\scalebox{0.9}{
\begin{tabular}{cccccccccccccccccccccc}
\toprule
    {\textbf{Metric}} &
	{\textbf{K}} &
	{\textbf{DCGAN}} &
	{\textbf{DCGAN + IM}} \\
\midrule
    \# Modes & 1/4 
    & $27.67 \pm 0.47$
    & $\mathbf{62.00 \pm 1.63}$
    \\
    \# Modes & 1/2 
    & $610.00 \pm 8.83$
    & $\mathbf{716.67 \pm 1.25}$ \\
    
    \midrule
    $\mathcal{D}_{\text{KL}}(p \vert\vert q)$ & 1/4 
    & $5.44 \pm 0.01$
    & $\mathbf{4.68 \pm 0.01}$ \\
    $\mathcal{D}_{\text{KL}}(p \vert\vert q)$ & 1/2 
    & $1.98 \pm 0.01$
    & $\mathbf{1.64 \pm 0.01}$ \\ 

\bottomrule
\end{tabular}
}
\vspace{0.1cm}
\caption{Number of modes (higher is better) recovered by the generator on the Stacked MNIST dataset, where the maximum value is 1000; and KL divergence $\mathcal{D}_{\text{KL}}(p \vert\vert q)$ between the distribution of generated modes $p$ and the uniform distribution $q$, where lower is better. `+ IM'' refers to adding our proposed InfoMax-GAN objective.}
\label{mode_collapse_table}
\end{table}

Following settings in \cite{metz2016unrolled}, we re-implement the DCGAN \cite{radford2015unsupervised} in \cite{metz2016unrolled} and evaluate its ability in recovering all 1000 modes of the Stacked MNIST dataset \cite{metz2016unrolled}, composed by randomly stacking 3 grayscale MNIST \cite{lecun1998mnist} digits into an RGB image, resulting in 1000 possible modes. We use a pre-trained MNIST classifier to classify each color channel of a generated image, and the model is said to recover 1 mode if it generates at least 1 image for that mode. We similarly set $K \in \{\frac{1}{4}, \frac{1}{2}\}$, where $K$ indicates the size of the discriminator relative to the generator.
Intuitively, the smaller $K$ is, the easier it is for the generator to fool the discriminator with just a few modes, resulting in less modes recovered.
Furthermore, we compute the KL divergence $\mathcal{D}_{\text{KL}}(p \vert\vert q)$ between the generated mode distribution $p$ and optimal uniform distribution of the modes $q$.
We see from Table \ref{mode_collapse_table} that our method helps to recover more modes for all $K$, with the recovered distribution having a consistently lower KL divergence with the ideal uniform distribution as a result.

\subsection{Additional Ablation Studies}
\label{appendix: ablation}
In this section, we analyze the impact of our framework design choices and their performance impact.

\begin{figure*}
\centering
\begin{subfigure}{.5\textwidth}
  \centering
    \includegraphics[width=\linewidth]{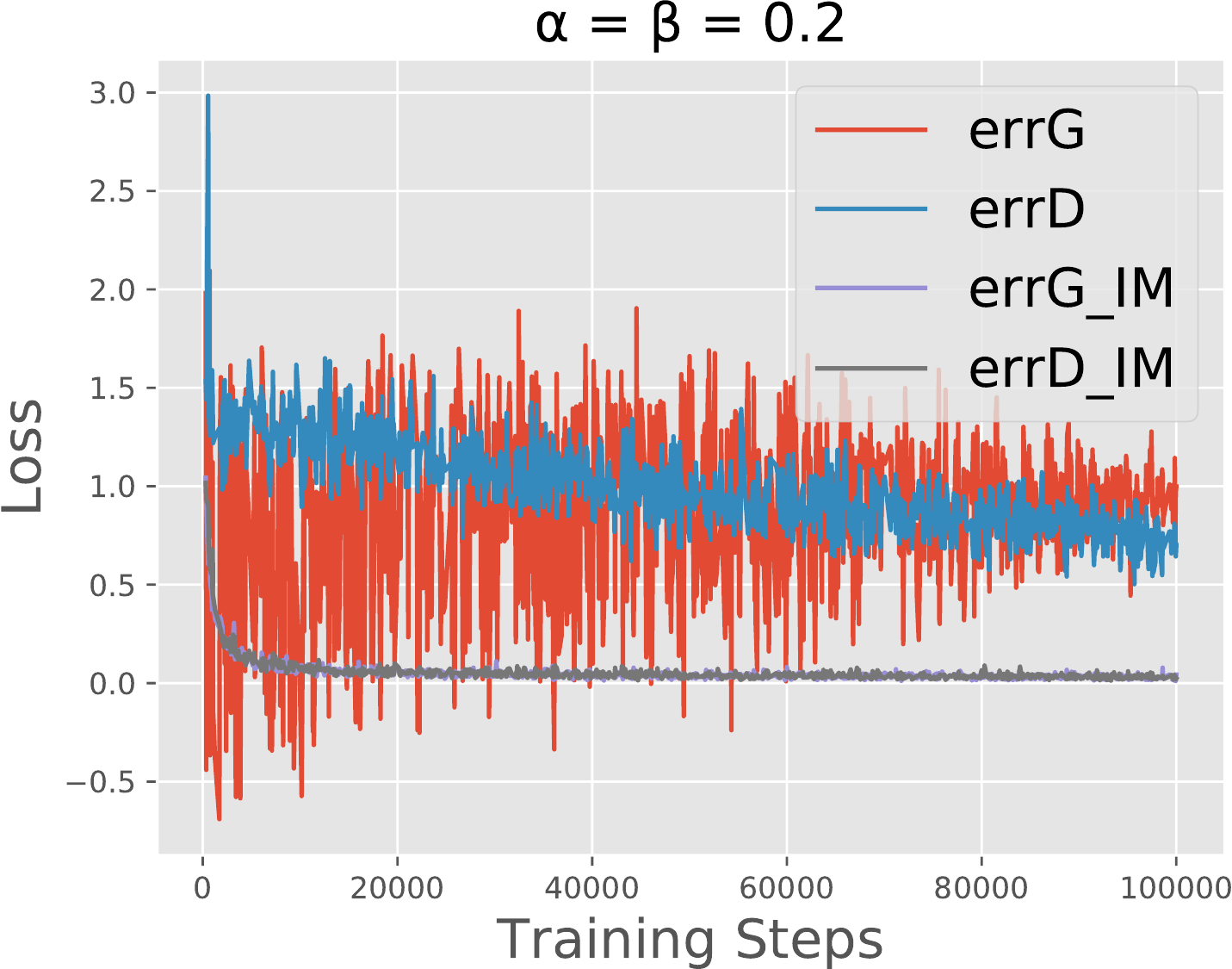}
\end{subfigure}%
\begin{subfigure}{.5\textwidth}
  \centering
    \includegraphics[width=\linewidth]{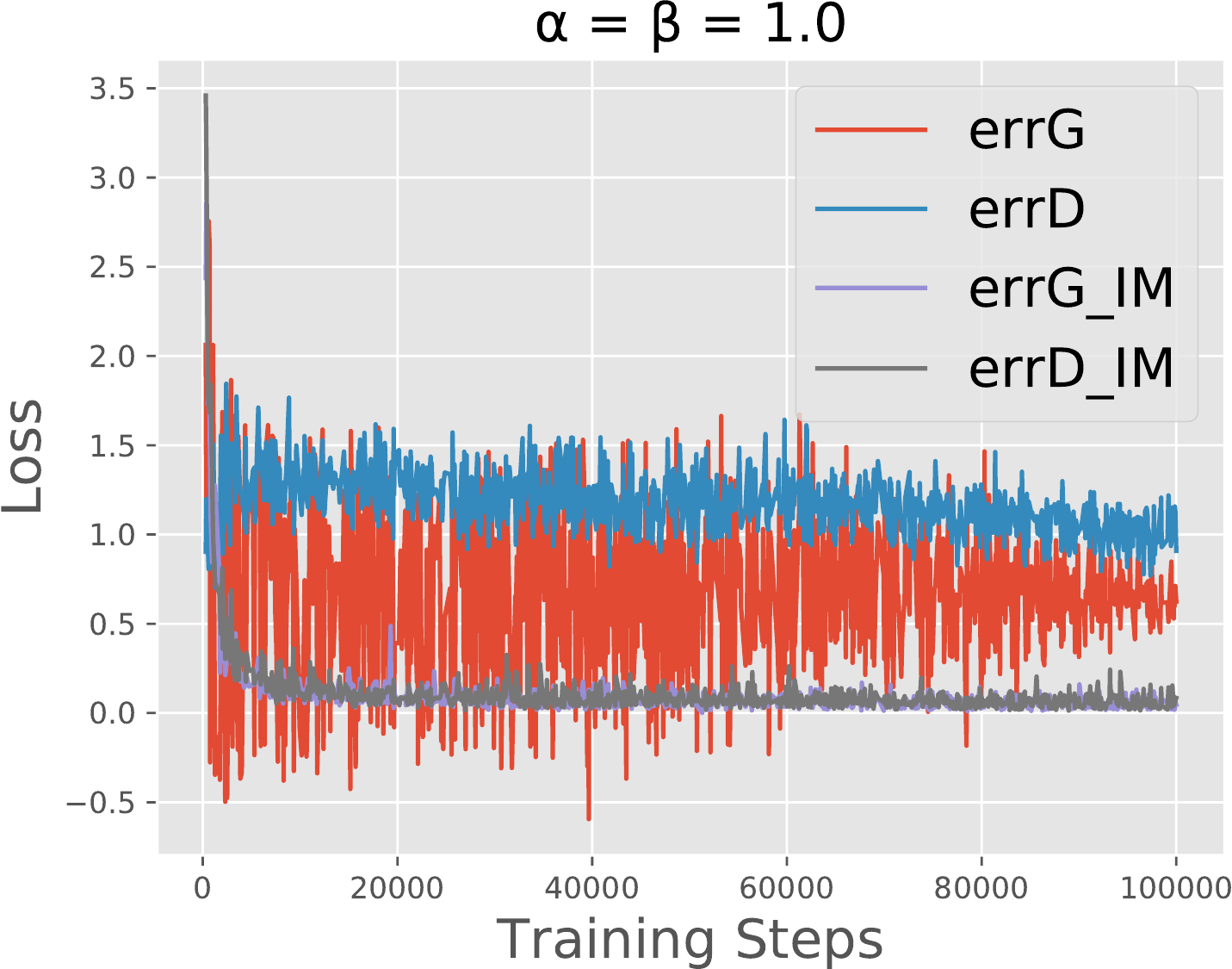}
\end{subfigure}
\caption{We show that the InfoMax objective loss decays very quickly regardless of the choice of scale for both $\alpha$ and $\beta$. \texttt{errD} and \texttt{errG} represents the GAN losses for the discriminator and generator respectively, and similarly, \texttt{errD\_IM} and \texttt{errG\_IM} represents the InfoMax objective losses for the discriminator and generator respectively.}
\label{gan_vs_infomax_loss_curve}
\end{figure*}

\paragraph{Relative Scale of Objective}
\label{gan_vs_infomax}
From Figure \ref{gan_vs_infomax_loss_curve}, we see in both our chosen hyperparameters of $\alpha=\beta=0.2$ and the other extreme of $\alpha=\beta=1.0$, the InfoMax objective loss decays very quickly relative to the GAN loss. In practice, we found that $\alpha=\beta=0.2$ performs better, which could be attributed to the relative magnitude of the InfoMax objective loss at the start of the training. When $\alpha=\beta=0.2$, the scales of the GAN and InfoMax objective losses are approximately equal initially. We highlight this is the same loss scaling principle applied in \cite{chen2016infogan}.

\paragraph{Position of feature maps} While we have chosen the local and global features to be the penultimate and final features of the discriminator encoder respectively, we examine the effect of alternative designs. For clarity, we note there is only \textit{one} global feature vector, which is the final feature output of the encoder. Correspondingly, our design can be called local-global, and other designs involving extracting intermediate local feature maps $C_{\psi, k}(x), 1 \leq k \leq n$ can be described as local-local. However, in practice, our original design of local-global is the only feasible option compared to local-local option, mainly due to the memory consumption: for any two feature maps of spatial size $M_1 \times M_1$ and $M_2 \times M_2$ respectively, we have the space complexity as $O(NM_1^2M_2^2R)$ for batch size $N$ and RKHS $R$. Fixing the first feature map size, the local-local approach has space complexity growing quadratically on the second feature map size $M_2$, which is in turn dependent on the image resolution. On the other hand, the local-global approach effectively sets $M_2=1$, which dramatically reduces memory consumption. 

In fact, in practice, we found the local-local approach cannot scale to datasets above $32 \times 32$ resolution as it would exceed 11GB for a single GPU. To still test this approach on the $32 \times 32$ resolution CIFAR-10 dataset, we reduce the memory consumption by randomly sampling only half of local spatial vectors from each feature map. Even so, the memory consumption is approximately 7 times of the local-global approach, making it highly memory intensive. In contrast, the local-global approach scales for even high resolution (e.g. $128 \times 128$) datasets and takes only a small portion of the memory size compared to the GAN models. Importantly, the local-local approach worsens FID by \textbf{3.1 points} from $17.14 \pm 0.20$ to $20.20 \pm 0.05$. Thus, this ablation study show that in practice, our current design is the most optimal for achieving both performance and memory consumption gains.

\paragraph{Effect of spectral normalizing critic} Interestingly, using spectral normalisation for the InfoMax-GAN critic networks leads to FID improvements. On CIFAR-10, using spectral normalisation for these critic networks improved FID by \textbf{1.5 points} from $18.67 \pm 0.25$ to $17.14 \pm 0.20$. We conjecture this could be related to the Wasserstein Dependency Measure \cite{ozair2019wasserstein}, a variant of mutual information which replaces the KL divergence term with Wasserstein distance, as measured using encoders from the class of 1-Lipschitz functions. However, in contrast to this work, our method enforces 1-Lipschitzness of the encoder using spectral normalization rather than gradient penalty. A theoretical treatment of this relationship is beyond the scope of this paper, which we leave as future work.

\subsection{Generated Image Samples}
\label{appendix B: generated_images}
\begin{figure*}[ht!]
\centering
\begin{subfigure}{.33\textwidth}
  \centering
    \includegraphics[width=\linewidth]{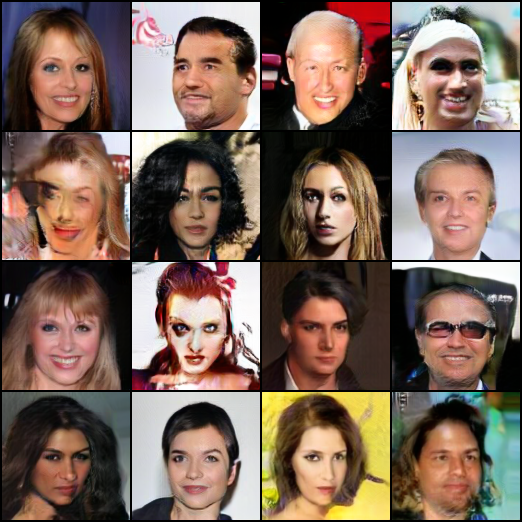}
    \caption{}
  \label{celeba_comp:sngan}
\end{subfigure}
\begin{subfigure}{.33\textwidth}
  \centering
    \includegraphics[width=\linewidth]{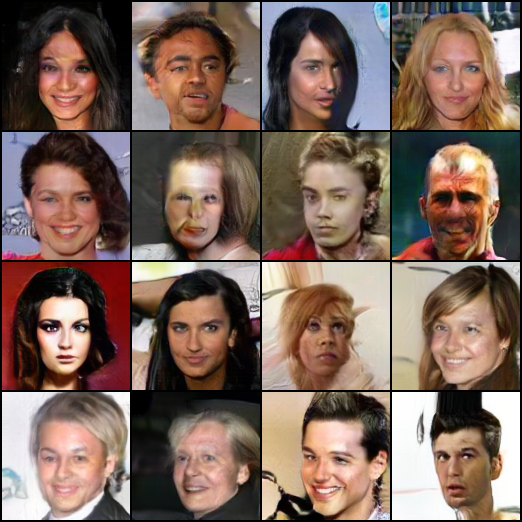}
    \caption{}
  \label{celeba_comp:ssgan}
\end{subfigure}
\begin{subfigure}{.33\textwidth}
  \centering
    \includegraphics[width=\linewidth]{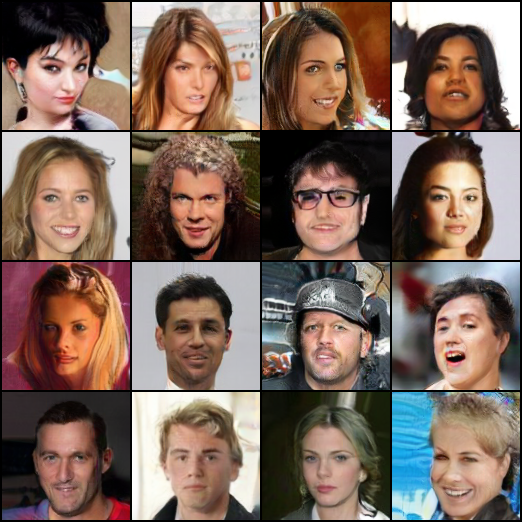}
    \caption{}
  \label{celeba_comp:infomax_gan}
\end{subfigure}
\caption{Generated CelebA images at $128 \times 128$ resolution for (a) SNGAN, (b) SSGAN, and (c) InfoMax-GAN. In general, we observe InfoMax-GAN generated images have less visual artifacts in both the background and the facial attributes. We note these images are randomly generated and non-cherry picked.}
\label{celeba_comp}
\end{figure*}

In Figure \ref{celeba_comp}, we show generated images at $128 \times 128$ resolution for CelebA. In general, we observe that images generated by InfoMax-GAN have less visual artifacts for both the background and facial attributes, with even attributes like spectacles and caps generated. In contrast, both SNGAN and SSGAN generated images tend to have more severe background artifacts, with certain prominent facial features like eyes and noses not well blended together. This blending problem is more commonly seen in SSGAN generated images, which may explain its worse FID performance compared to both SNGAN and InfoMax-GAN. We further provide image samples randomly generated for all datasets in Appendix \ref{appendix B: generated_images}. 

\begin{figure*}
\centering
  \caption{Randomly sampled and non-cherry picked images for SNGAN (left) and InfoMax-GAN (right) for CIFAR-10, CIFAR-100, and STL-10.} 
  \label{sup_images_a}
    \begin{subfigure}[t]{\textwidth}
        \centering
        \includegraphics[width=0.4\linewidth]{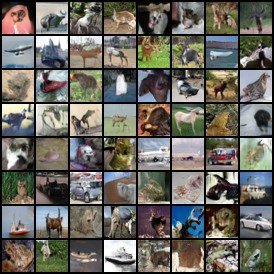}
        \includegraphics[width=0.4\linewidth]{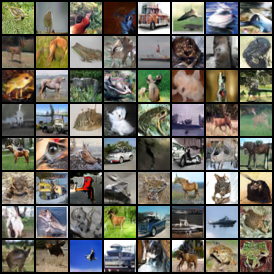}
        \caption{CIFAR-10.}
    \end{subfigure}
    \begin{subfigure}[t]{\textwidth}
        \centering
        \includegraphics[width=0.4\linewidth]{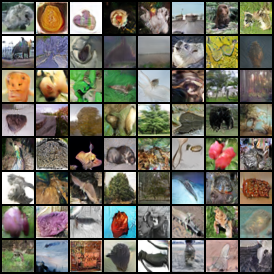}
        \includegraphics[width=0.4\linewidth]{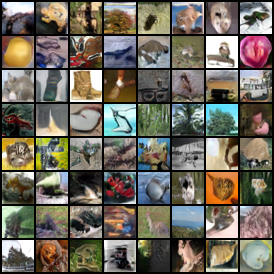}
        \caption{CIFAR-100.}
    \end{subfigure}    
    \begin{subfigure}[t]{\textwidth}
        \centering
        \includegraphics[width=0.4\linewidth]{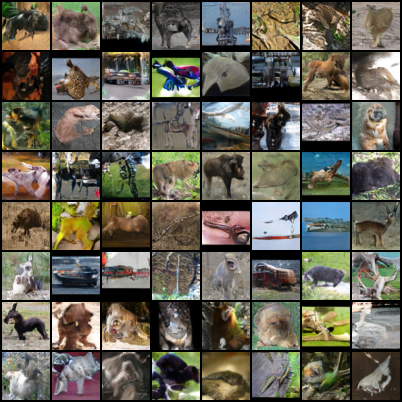}
        \includegraphics[width=0.4\linewidth]{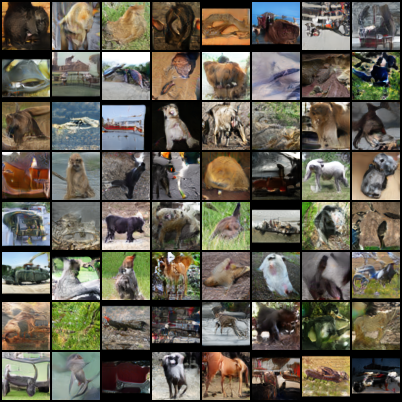}
        \caption{STL-10.}
    \end{subfigure}
\end{figure*}

\begin{figure*}
\centering
  \caption{Randomly sampled and non-cherry picked generated CelebA images for SNGAN (top) and InfoMax-GAN (bottom).} 
  \label{sup_images_b}
\begin{subfigure}{\textwidth}
    \centering
    \includegraphics[width=0.65\linewidth]{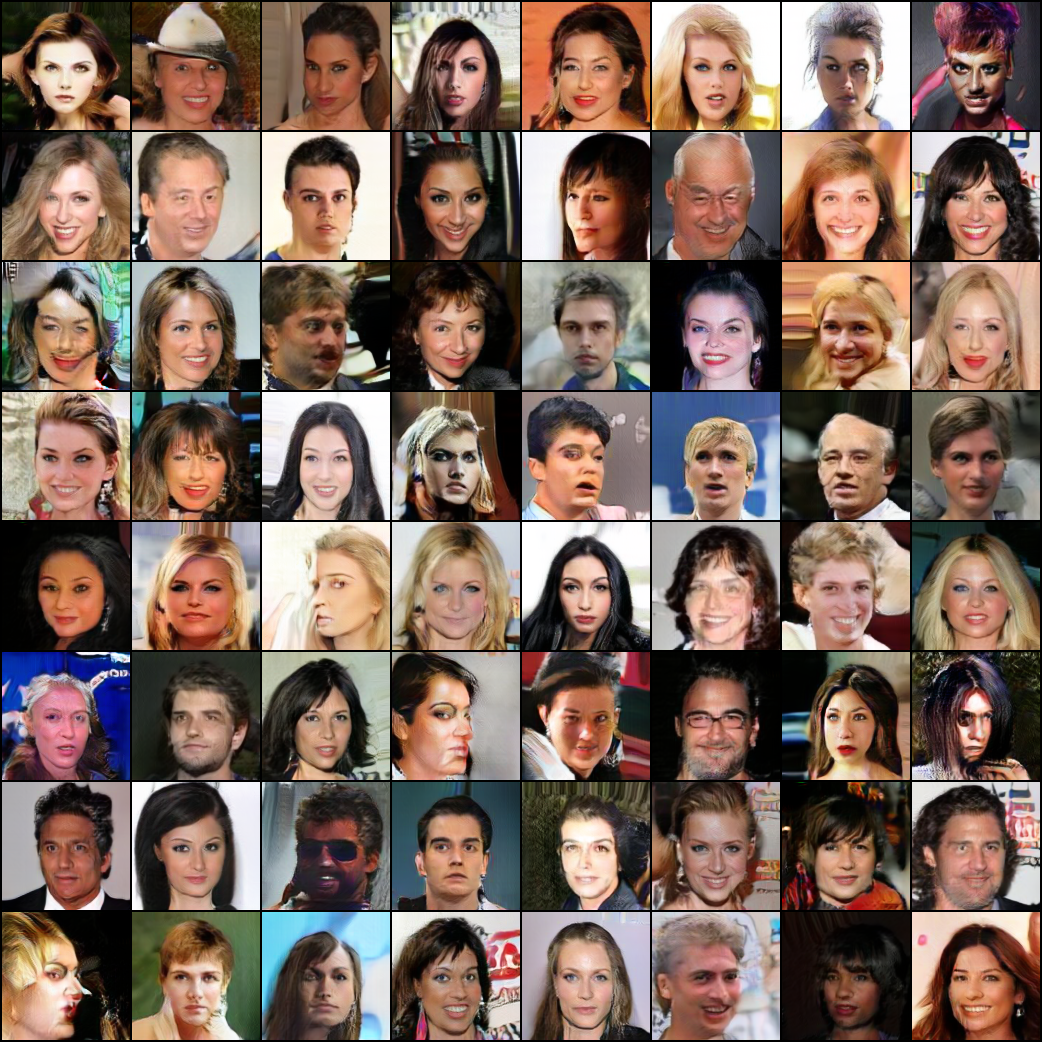}
    \vspace{0.5cm}
\end{subfigure}
\begin{subfigure}{\textwidth}
    \centering
    \includegraphics[width=0.65\linewidth]{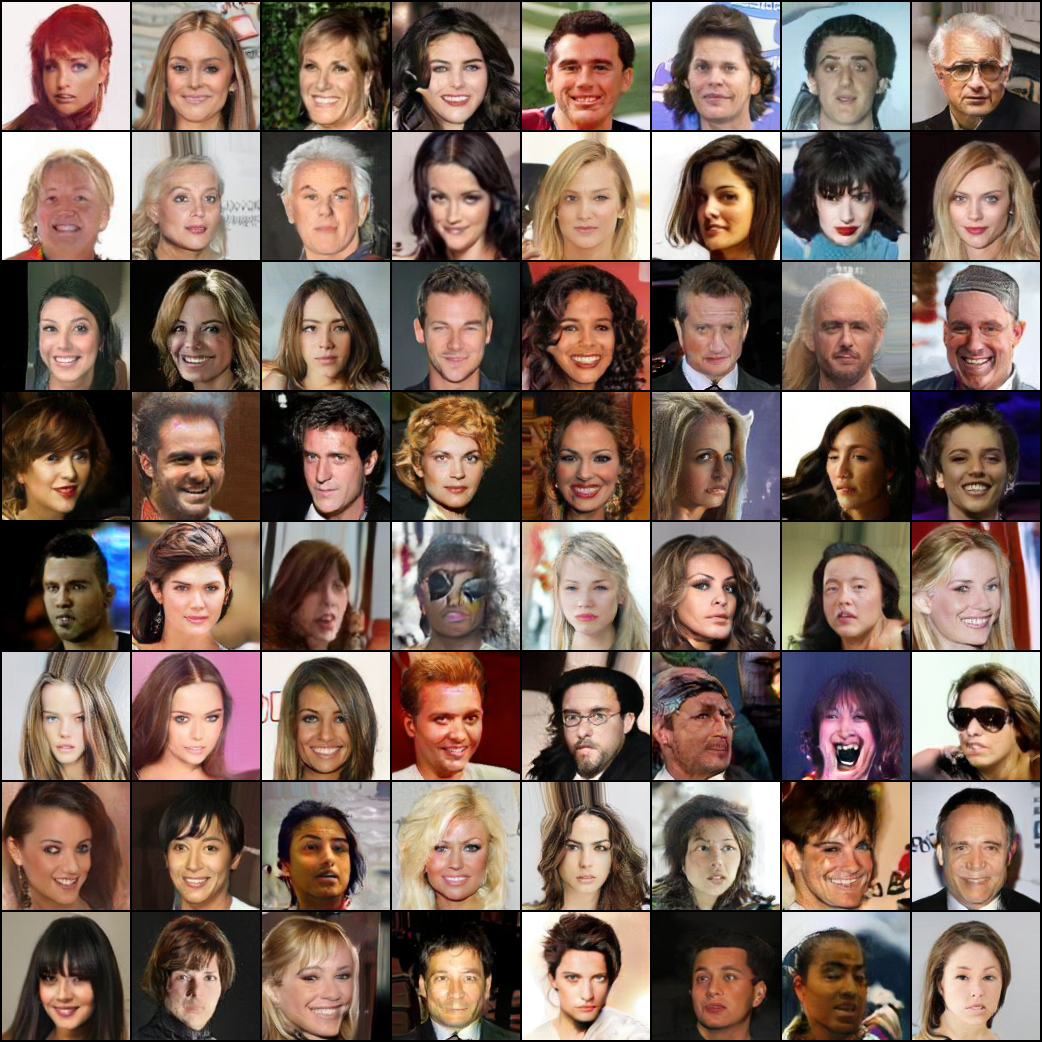}
\end{subfigure}
\end{figure*}

\begin{figure*}
\centering
  \caption{Randomly sampled and non-cherry picked generated ImageNet images for SNGAN (top) and InfoMax-GAN (bottom).} 
  \label{sup_images_c}
\begin{subfigure}{\textwidth}
    \centering
    \includegraphics[width=0.65\linewidth]{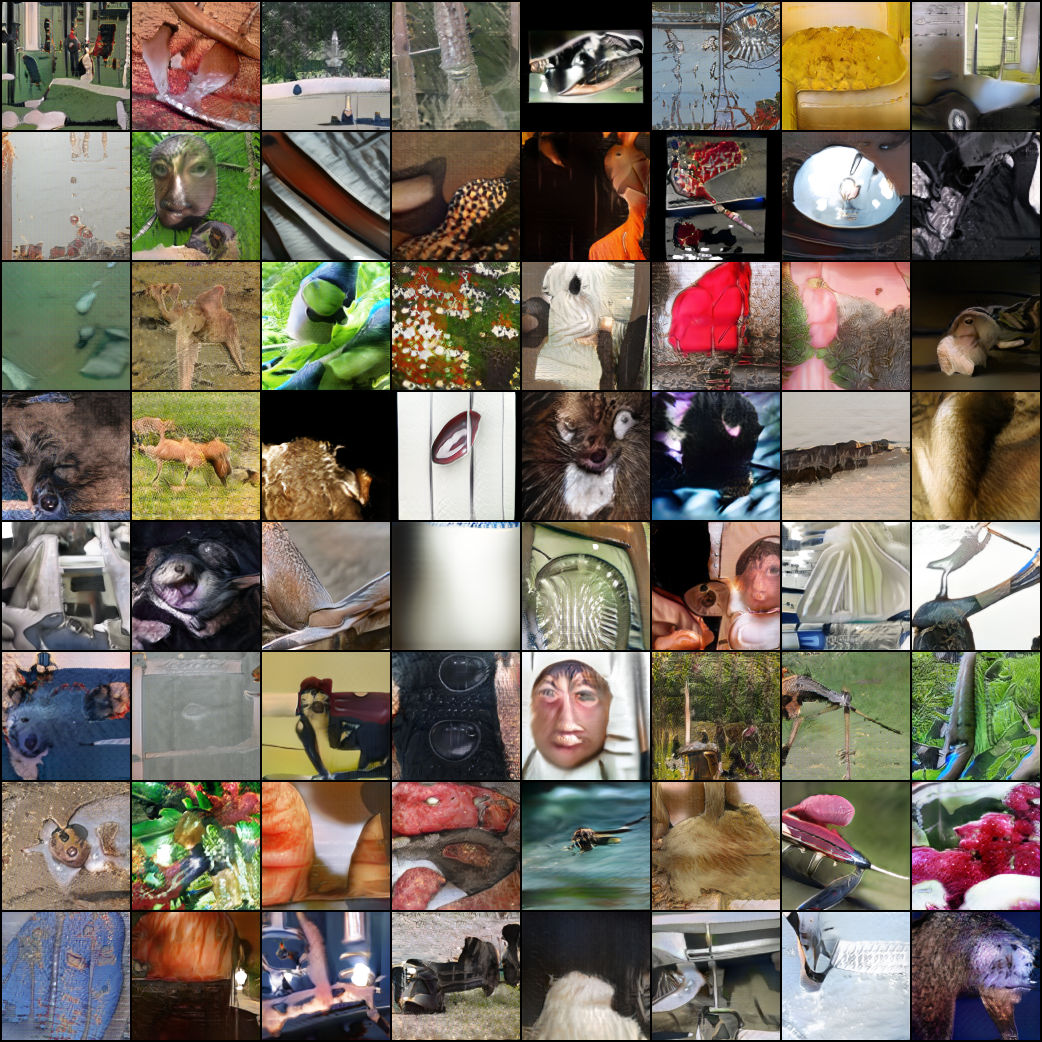}
    \vspace{0.5cm}
\end{subfigure}
\begin{subfigure}{\textwidth}
    \centering
    \includegraphics[width=0.65\linewidth]{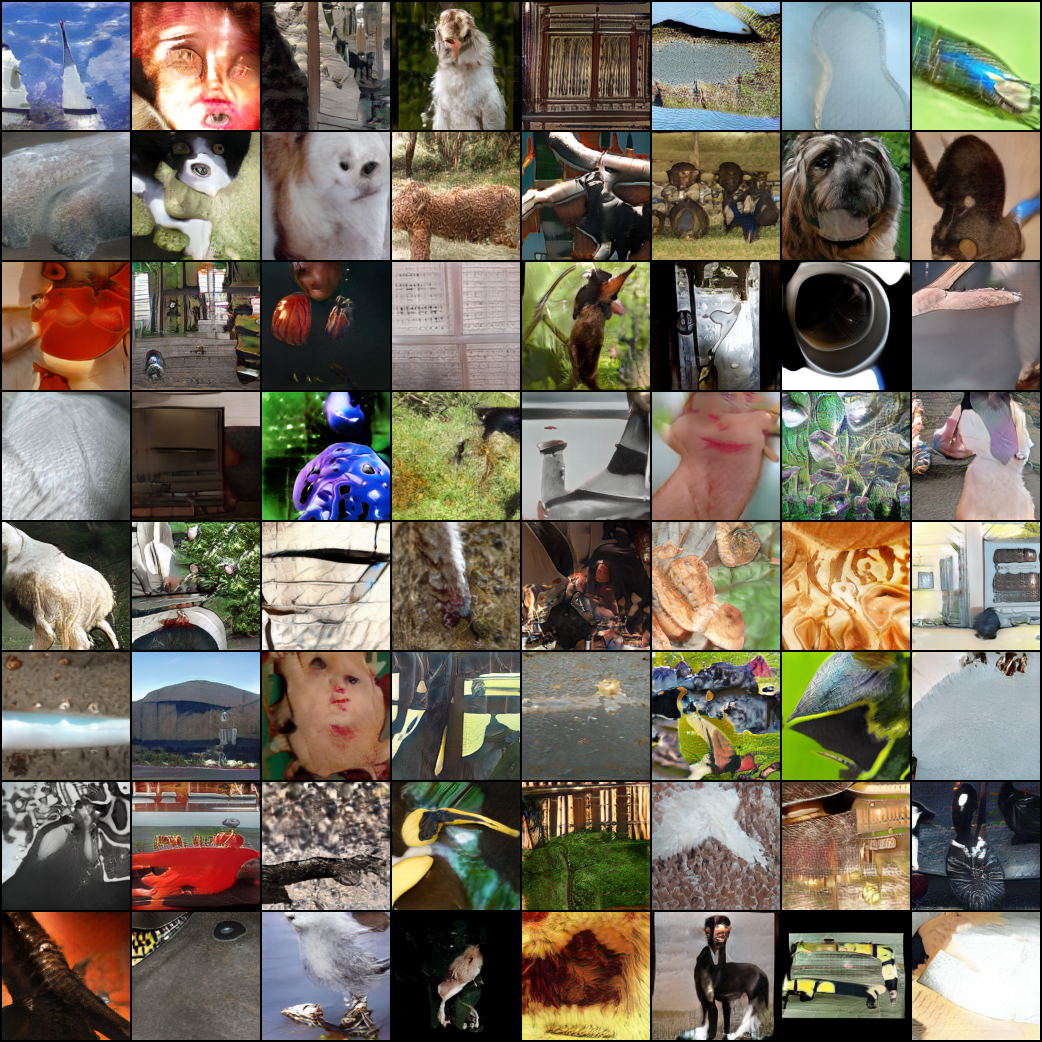}
\end{subfigure}
\end{figure*}

For further qualitative comparisons, we present randomly sampled, non-cherry picked images generated by SNGAN and InfoMax-GAN for all datasets in Figures \ref{sup_images_a}, \ref{sup_images_b} and \ref{sup_images_c}. We qualitatively observe that the images are more diverse and have sharper shapes after the use of an InfoMax objective.

\section{InfoGAN Comparison}
\label{appendix: infogan_comparison}
\begin{table*}
\centering
\begin{tabular}{@{}cccc@{}}
\toprule
\textbf{Work} & \textbf{Target Outcome} & \textbf{MI Objective} & \textbf{\begin{tabular}[c]{@{}c@{}}MI \\ Approximation\\  Technique\end{tabular}} \\ \midrule
InfoGAN \cite{chen2016infogan} & \begin{tabular}[c]{@{}c@{}}Disentangled representation learning \\ by using an input encoding $c$ \\ to the generator to control its output.\end{tabular} & $\mathcal{I}(c; G(z, c))$ & \begin{tabular}[c]{@{}c@{}}Variational\\  Information\\  Maximization \cite{barber2003algorithm}\end{tabular} \\
\midrule
InfoMax-GAN (ours) & \begin{tabular}[c]{@{}c@{}}Improve image synthesis by reducing\\ catastrophic forgetting of discriminator\\  and mode collapse of generator.\end{tabular} & $\mathcal{I}(C_\psi(X); E_\psi(X))$ & InfoNCE \cite{oord2018representation} Task \\ \bottomrule
\end{tabular}
\vspace{0.1cm}
\caption{Comprehensive differences with InfoGAN. Our work mainly differs in the intended outcome, the objective to meet the outcome, and the approximation technique needed to solve the objective.}
\label{infogan_comparison_table}
\end{table*}

For clarity and disambiguity, Table \ref{infogan_comparison_table} illustrates the differences in our work with InfoGAN. Our works have different focuses: InfoGAN focuses on learning disentanglements in image generation, while we focus on improving image synthesis as a whole.

\section{Model Architectures}
\label{appendix:B}
We detail the exact GAN architectures used for all datasets in Tables \ref{32_archs}, \ref{48_archs}, \ref{128_archs}. We also detail the architectures for projecting the local and global features to a higher dimensional RKHS for solving the InfoNCE task in Table \ref{InfoNCE_archs}.

\begin{table*}[!htb]
    \caption{Network architectures for the CIFAR-10 and CIFAR-100 datasets, which follows exact settings in \cite{miyato2018spectral}.}
    \begin{subtable}{.5\linewidth}
      \centering
        \caption{Generator}
        \begin{tabular}{c}
            \toprule
            \midrule
            $z \in \mathbb{R}^{128} \sim \mathcal{N}(0, 1)$ \\
            \midrule
            Linear, $4 \times 4 \times 256$ \\
            \midrule
            ResBlock up 256 \\
            \midrule
            ResBlock up 256 \\
            \midrule
            ResBlock up 256 \\            
            \midrule
            BN; ReLU; $3\times3$ conv, 3; Tanh \\
            \midrule
            \bottomrule
        \end{tabular}
    \end{subtable}
    \begin{subtable}{.5\linewidth}
      \centering
        \caption{Discriminator}
        \begin{tabular}{c}
            \toprule
            \midrule
            RGB image $x \in \mathbb{R}^{32 \times 32 \times 3}$ \\
            \midrule
            ResBlock down 128 \\
            \midrule
            ResBlock down 128 \\
            \midrule
            ResBlock 128 $\rightarrow$ Local Features \\
            \midrule
            ResBlock 128 \\
            \midrule
            ReLU \\
            \midrule
            Global Sum Pooling $\rightarrow$ Global Features \\
            \midrule
            Linear $\rightarrow$ 1 \\
            \midrule
            \bottomrule
        \end{tabular}
    \end{subtable}
    \begin{subtable}{.5\linewidth}
      \centering
        \vspace{0.5cm}
        \caption{Self-supervised Discriminator}
        \begin{tabular}{c}
            \toprule
            \midrule
            RGB image $x \in \mathbb{R}^{32 \times 32 \times 3}$ \\
            \midrule
            ResBlock down 128 \\
            \midrule
            ResBlock down 128 \\
            \midrule
            ResBlock 128 $\rightarrow$ Local Features \\
            \midrule
            ResBlock 128 \\
            \midrule
            ReLU \\
            \midrule
            Global Sum Pooling $\rightarrow$ Global Features \\
            \midrule
            Linear  $\rightarrow$ 1; Linear $\rightarrow$ 4 \\
            \midrule
            \bottomrule
        \end{tabular}
    \end{subtable}     
    \label{32_archs}
\end{table*}

\begin{table*}[!htb]
    \caption{Network architectures for the STL-10 dataset, which follows exact settings in \cite{miyato2018spectral}.}
    \begin{subtable}{.5\linewidth}
      \centering
        \caption{Generator}
        \begin{tabular}{c}
            \toprule
            \midrule
            $z \in \mathbb{R}^{128} \sim \mathcal{N}(0, 1)$ \\
            \midrule
            Linear, $6 \times 6 \times 512$ \\
            \midrule
            ResBlock up 256 \\
            \midrule
            ResBlock up 128 \\
            \midrule
            ResBlock up 64 \\            
            \midrule
            BN; ReLU; $3\times3$ conv, 3; Tanh \\
            \midrule
            \bottomrule
        \end{tabular}
    \end{subtable}
    \begin{subtable}{.5\linewidth}
      \centering
        \caption{Discriminator}
        \begin{tabular}{c}
            \toprule
            \midrule
            RGB image $x \in \mathbb{R}^{48 \times 48 \times 3}$ \\
            \midrule
            ResBlock down 64 \\
            \midrule
            ResBlock down 128 \\
            \midrule
            ResBlock down 256 \\
            \midrule
            ResBlock down 512 $\rightarrow$ Local Features \\
            \midrule
            ResBlock 1024 \\
            \midrule
            ReLU \\
            \midrule
            Global Sum Pooling $\rightarrow$ Global Features \\
            \midrule
            Linear $\rightarrow$ 1 \\
            \midrule
            \bottomrule
        \end{tabular}
    \end{subtable}
    \begin{subtable}{.5\linewidth}
      \centering
        \vspace{0.5cm}
        \caption{Self-supervised Discriminator}
        \begin{tabular}{c}
            \toprule
            \midrule
            RGB image $x \in \mathbb{R}^{48 \times 48 \times 3}$ \\
            \midrule
            ResBlock down 64 \\
            \midrule
            ResBlock down 128 \\
            \midrule
            ResBlock down 256 \\
            \midrule
            ResBlock down 512 $\rightarrow$ Local Features \\
            \midrule
            ResBlock 1024 \\
            \midrule
            ReLU \\
            \midrule
            Global Sum Pooling $\rightarrow$ Global Features \\
            \midrule
            Linear  $\rightarrow$ 1; Linear $\rightarrow$ 4 \\
            \midrule
            \bottomrule
        \end{tabular}
    \end{subtable}     
    \label{48_archs}
\end{table*}

\begin{table*}[!htb]
    \caption{Network architectures for the CelebA and ImageNet datasets. This follows the exact settings in the official SNGAN code \cite{pfnet}.}
    \begin{subtable}{.5\linewidth}
      \centering
        \caption{Generator}
        \begin{tabular}{c}
            \toprule
            \midrule
            $z \in \mathbb{R}^{128} \sim \mathcal{N}(0, 1)$ \\
            \midrule
            Linear, $4 \times 4 \times 1024$ \\
            \midrule
            ResBlock up 1024 \\
            \midrule
            ResBlock up 512 \\
            \midrule
            ResBlock up 256 \\
            \midrule
            ResBlock up 128 \\
            \midrule
            ResBlock up 64 \\            
            \midrule
            BN; ReLU; $3\times3$ conv, 3; Tanh \\
            \midrule
            \bottomrule
        \end{tabular}
    \end{subtable}
    \begin{subtable}{.5\linewidth}
      \centering
        \caption{Discriminator}
        \begin{tabular}{c}
            \toprule
            \midrule
            RGB image $x \in \mathbb{R}^{128 \times 128 \times 3}$ \\
            \midrule
            ResBlock down 64 \\
            \midrule
            ResBlock down 128 \\
            \midrule
            ResBlock down 256 \\
            \midrule
            ResBlock down 512 $\rightarrow$ Local Features \\
            \midrule
            ResBlock down 1024 \\
            \midrule
            ResBlock 1024 \\
            \midrule
            ReLU \\
            \midrule
            Global Sum Pooling $\rightarrow$ Global Features \\
            \midrule
            Linear $\rightarrow$ 1 \\
            \midrule
            \bottomrule
        \end{tabular}
    \end{subtable}
    \begin{subtable}{.5\linewidth}
      \centering
        \vspace{0.5cm}
        \caption{Self-supervised Discriminator}
        \begin{tabular}{c}
            \toprule
            \midrule
            RGB image $x \in \mathbb{R}^{128 \times 128 \times 3}$ \\
            \midrule
            ResBlock down 64 \\
            \midrule
            ResBlock down 128 \\
            \midrule
            ResBlock down 256 \\
            \midrule
            ResBlock down 512 $\rightarrow$ Local Features \\
            \midrule
            ResBlock down 1024 \\
            \midrule
            ResBlock 1024 \\
            \midrule
            ReLU \\
            \midrule
            Global Sum Pooling $\rightarrow$ Global Features \\
            \midrule
            Linear  $\rightarrow$ 1; Linear $\rightarrow$ 4 \\
            \midrule
            \bottomrule
        \end{tabular}
    \end{subtable}  
    \label{128_archs}
\end{table*}

\begin{table*}[!htb]
    \caption{InfoNCE projection architectures, which follow what were proposed in \cite{hjelm2018learning}. In practice, we extract the local features and global features from the penultimate and final residual blocks of the discriminator respectively. This decides the corresponding values of feature depth $K$.}
    \begin{subtable}{.5\linewidth}
      \centering
        \caption{Local features projection architecture.}
        \begin{tabular}{c}
            \toprule
            \midrule
            $1 \times 1$ Conv, $K$; $1 \times 1$ Conv, $R$ $\rightarrow$ Shortcut \\
            \midrule
            ReLU \\
            \midrule
            $1 \times 1$ Conv, $R$ + Shortcut \\
            \midrule
            \bottomrule
        \end{tabular}
    \end{subtable}
    \begin{subtable}{.5\linewidth}
      \centering
        \caption{Global features projection architecture.}
        \begin{tabular}{c}
            \toprule
            \midrule
            Linear $\rightarrow K$; Linear $\rightarrow R \rightarrow$ Shortcut \\
            \midrule
            ReLU \\
            \midrule
            $1 \times 1$ Conv, $R$ + Shortcut \\
            \midrule
            \bottomrule
        \end{tabular}
    \end{subtable}
    \label{InfoNCE_archs}
\end{table*}

\end{document}